\documentclass[journal=jcisd8,manuscript=article]{achemso}

\usepackage{geometry}
\usepackage{setspace}

\usepackage{tabularx}
\usepackage{enumitem}
\usepackage{amsmath}
\usepackage{amssymb}
\usepackage{amsthm}
\usepackage{bm}
\usepackage{mathtools}

\usepackage{graphicx}
\usepackage{subcaption}
\usepackage{float}

\usepackage{booktabs}
\usepackage{multirow}
\usepackage{makecell}
\usepackage{array}
\usepackage{tabularx}
\usepackage{adjustbox}
\usepackage{siunitx}
\usepackage{threeparttable}

\usepackage[table]{xcolor}

\usepackage{algorithm}
\usepackage{algpseudocode}

\usepackage[version=4]{mhchem}   

\newfloat{scheme}{htbp}{los}
\floatname{scheme}{Scheme}
\newfloat{graph}{htbp}{loh}

\usepackage{hyperref}
\usepackage{cleveref}

\usepackage{comment}
\usepackage{kotex}

\author{Bumju Kwak}
\affiliation{Independent researcher, Seoul, Republic of Korea}

\author{Jeonghee Jo}
\email{jade.jeonghee.jo@gmail.com}
\affiliation{Independent researcher, Seoul, Republic of Korea}

\title{Hessian-Informed Conformation Augmentation for Potential Energy Surface Curvature in Machine Learning Interatomic Potentials}

\begin{document}

\maketitle
\begin{abstract}
While machine-learning interatomic potentials (MLIPs) have successfully learned potential energy surfaces (PES) and atomic forces, many practical applications, such as vibrational analysis and transition state search, rely heavily on the PES Hessian. Yet standard MLIPs are trained on energy and forces only, and using the Hessian as an additional training objective requires backpropagation whose memory cost grows quadratically with the number of atoms. To address these limitations, we propose two Hessian-derived data augmentation schemes: isotropic Gaussian displacement (\textbf{UniAug}) and normal mode-weighted displacement (\textbf{ModeAug}). Both methods utilize simple Taylor expansions, generating physically grounded molecular structures with highly accurate energy and force labels. By preserving the underlying MLIP architecture and training procedure, this approach contributes to improve energy and force predictions in various benchmarks. Comprehensive evaluations across non-equilibrium and equilibrium datasets demonstrate that our approach enhances model accuracy where reference forces are large and training data are limited.

\end{abstract}




\section{Introduction}
Energies and forces of molecules have been obtained from first principles calculations such as DFT, and machine learning interatomic potentials (MLIPs), which learn DFT energies and forces with neural networks, are rapidly replacing these calculations~\citep{behler2007generalized, bartok2010gaussian, behler2021four, unke2021machine}. MLIPs predict molecular energies and forces with high accuracy across diverse domains and have greatly advanced molecular dynamics simulations~\citep{chanussot2021open, poltavsky2021machine, unke2024biomolecular}.

However, energy and force labels provide only the value and gradient of the PES at a single structure. The information that determines properties such as vibrational modes and transition states resides in the curvature of the PES, the Hessian~\citep{yuan2024analytical, fang2024phonon}, which is not sufficiently constrained by the energy and force of a single structure. \citet{fu2023forces} showed that even state-of-the-art machine learning force field models may fail to reproduce dynamical properties in molecular dynamics simulations despite achieving low force errors, and similar results have been reported elsewhere~\citep{cui2025large, rodriguez2025does}. This implies that existing reference data carry only limited curvature information.

Direct use of the Hessian in MLIPs remains at a relatively early stage~\citep{amin2025distilling, rodriguez2025does}. Large scale datasets that include Hessians have only recently been released, namely HORM~\citep{cui2025large}, which covers nonequilibrium structures including transition states, and HessianQM9~\citep{williams2025hessian}, which focuses only on equilibrium structures. Training on the Hessian as a loss, however, makes the backpropagation graph grow quadratically with the number of atoms, which makes training inefficient~\citep{rodriguez2026projected, nils2025beyond}. To solve this problem, \citet{cui2025large, amin2025distilling} used row sampling to achieve linear growth, other approaches predict forces directly and obtain the Hessian as their derivative~\citep{gasteiger2021gemnet, liao2024equiformerv2}, and another predicts the Hessian without differentiation through an equivariant readout head~\citep{burger2025shoot}. All of these approaches, however, treat the Hessian as an explicit supervision target, requiring an additional loss and modifying a standard MLIP architecture.

Alternatively, we propose a data augmentation strategy that uses the Hessian of an original molecule to generate new training structures, rather than treating it as a supervision target. This augmentation generates physically grounded molecular structures by local perturbation around an original structure, and energies from a second order Taylor expansion and forces from a first order expansion serve as labels. Because under the Born–Oppenheimer (BO) approximation the PES is smooth with respect to atomic coordinates, higher order terms can be assumed small enough to be ignored in a local region, allowing the Taylor expansion to yield reliable energy and force labels. By this augmentation, local information around the molecular PES can be indirectly learned during the training process without any additional training objective.

Molecular data augmentation has been explored in various forms~\citep{cooper2020efficient, godwin2022simple, yeu2025scalable, kulichenko2024datagen}, but existing schemes either displace structures without reference to the local PES, which leaves the original labels inaccurate for the perturbed structures~\citep{gibson2022data}, or obtain valid labels only by running additional \textit{ab initio} calculations on the generated structures~\citep{smith2017ani}. Meanwhile, physics-constrained data augmentation has also been proposed for reactive MLIPs. \citet{f2025improving} augmented training data using a Morse potential and a reference MLIP, which improved the predictive performance and stability of models for bond dissociation structures. Our strategy addresses both limitations by providing valid labels for the displaced structures from the Hessian of the original molecule.

This design leads to several advantages in applicability and scalability. First, this scheme changes neither the objective nor the model architecture and therefore applies directly to existing MLIP architectures and training pipelines. Second, for the same number of training samples, the backpropagation memory cost equals that of energy and force training. Finally, different perturbations of a single structure also yield multiple structure and label pairs, which complement the structural diversity of the reference data.

Specifically, we propose two schemes according to the perturbation method, UniAug and ModeAug. UniAug applies an isotropic random displacement, whereas ModeAug applies a normal mode dependent perturbation. By reflecting the vibrational characteristics of molecules, the additional design in ModeAug can lead to smaller changes in energy and forces from the original structure compared to UniAug, even at larger displacements. We first verify that the Taylor-approximated energies of augmented structures provide a reliable approximation of reference DFT energies. We then compare energy and force predictions with and without augmentation on both non-equilibrium and equilibrium structures, and perform zero shot evaluation on dynamics and curvature recovery tasks. We find that the proposed augmentation is particularly effective for nonequilibrium structures, where atomic forces are distributed over larger values. Moreover, the performance improvement from augmentation is clearer than that from Hessian supervision when training data are limited.  Finally, we compare the backpropagation memory cost during training against Hessian supervision. Our contributions are summarized below.

\begin{itemize}
  \item We introduce two augmentation schemes that draw on the Hessian of a molecular structure, UniAug and ModeAug. Both methods utilize curvature information at the data level via Taylor expansions, preserving the underlying architecture’s physical constraints.
  \item We show that both schemes generate physically grounded structures and that the energies obtained from the Taylor expansion remain reliable at the dataset level. ModeAug keeps changes in energy and forces small at larger displacements.
  \item Across multiple benchmarks, we observe that both schemes are especially useful for nonequilibrium structures and in data scarce settings, while being computationally cheaper than Hessian supervision.
\end{itemize}

\begin{figure}[htbp]
  \centering
  \begin{subfigure}[t]{0.58\linewidth}
    \centering
    \includegraphics[width=\linewidth]{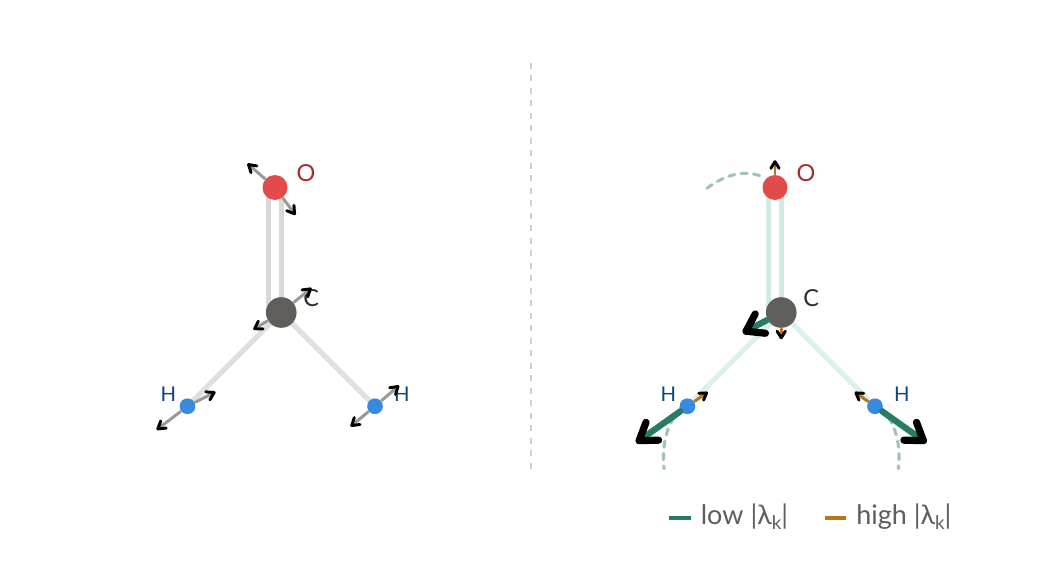}
    \caption{Overview of the two augmentation strategies.}
    \label{fig:aug_overview_schemes}
  \end{subfigure}
  \hfill
  \begin{subfigure}[t]{0.38\linewidth}
    \centering
    \includegraphics[width=\linewidth]{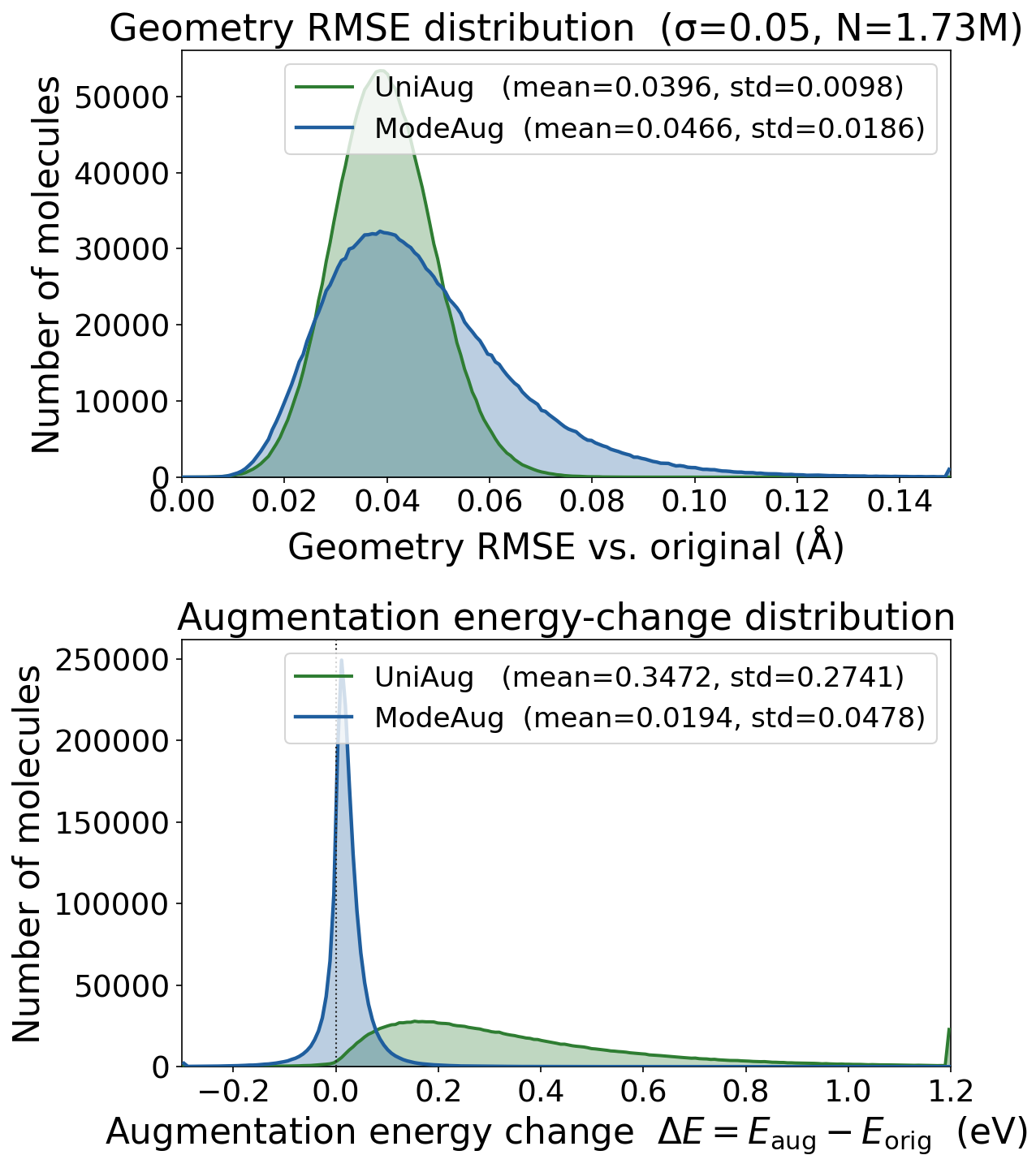}
    \caption{Distributions of geometry RMSE (top) and Taylor energy change ΔE (bottom) for UniAug and ModeAug.}
    \label{fig:aug_overview_rmse}
  \end{subfigure}
    \caption{Overview of the two augmentation schemes. (a) UniAug displaces a random subset of atoms isotropically in Cartesian space, ModeAug along the normal modes of the mass-weighted Hessian with amplitude weighted by $1/\sqrt{|\lambda_k|}$ ($\lambda_k$: eigenvalue of mode $k$). (b) Geometry RMSE and Taylor-label $\Delta E$ of the augmented structures on \texttt{ts1x-train} at $\sigma = 0.05$\,\AA.}
  \label{fig:aug_overview}
\end{figure}

\section{Background}

\subsection{Force prediction in MLIPs}



The force on atom $i$, $\mathbf{F}_i \in \mathbb{R}^3$, is the negative gradient of the energy with respect to its position $\mathbf{X}_i$, and MLIPs use forces alongside energies as a joint training target. Forces are predicted in one of two ways. The first is conservative force training~\citep{schutt2018schnet, gasteiger2020fast, schutt2021equivariant, batzner2022nequip, haghighatlari2022newtonnet, batatia2022mace, musaelian2023allegro}, in which the predicted force $\hat{\mathbf{F}}_i$ is obtained by automatic differentiation of the predicted energy $\hat{E}$, $\hat{\mathbf{F}}_i = -\nabla_{\mathbf{X}_i} \hat{E}$. The resulting force field is by construction the negative gradient of a scalar potential and is therefore conservative. This avoids the unphysical energy drift that non-conservative fields induce in molecular dynamics (MD) simulations. However, since the predicted force is itself obtained by differentiation, backpropagating the force loss requires second-order automatic differentiation.

The second is direct force training, in which the model predicts atomic forces through separate network outputs rather than obtaining them by differentiation~\citep{gasteiger2021gemnet, gasteiger2022gemnet, passaro2023reducing}. This accelerates training, reduces memory consumption, and removes the differentiability requirement that gradient-based models impose on the cutoff function~\citep{gasteiger2021gemnet}. Nevertheless, directly predicted force fields are not guaranteed to be conservative. The force Jacobian $\partial \mathbf{\hat{F}_i} / \partial \mathbf{X}_j$ is also generally asymmetric, and therefore does not correspond to a valid energy Hessian. Despite these limitations, direct force training has been reported to achieve lower force MAEs than automatic differentiation approaches on several large-scale benchmarks~\citep{gasteiger2021gemnet, liao2024equiformerv2}.

\subsection{Second-order Taylor approximation of the PES}
The Taylor series approximates a smooth function near a reference point using only local derivative information at that point. Under the Born--Oppenheimer approximation, the PES is inherently smooth, so such local expansions are well defined. Around a reference configuration $\mathbf{X}_0$, the energy at a displaced position $\mathbf{X}_0 + \boldsymbol{\delta}$ is approximated to second order as
\begin{equation}
E(\mathbf{X}_0 + \boldsymbol{\delta})
\approx E(\mathbf{X}_0)
- \mathbf{F}(\mathbf{X}_0)^\top \boldsymbol{\delta}
+ \frac{1}{2}\,\boldsymbol{\delta}^\top \mathbf{H} \,\boldsymbol{\delta},
\label{eq:pes_taylor}
\end{equation}

where $\mathbf{H} = \nabla^2 E(\mathbf{X}_0)$ is the molecular Hessian matrix, and $\mathbf{F}(\mathbf{X}_0) = -\nabla E(\mathbf{X}_0)$ denotes the reference force vector. 
Unlike the harmonic approximation around a minimum, the expansion retains the first-order term, which is non-zero for off-equilibrium structures.
Differentiating Eq.~\eqref{eq:pes_taylor} with respect to position gives the corresponding force vector:
\begin{equation}
\mathbf{F}(\mathbf{X}_0 + \boldsymbol{\delta})
= -\nabla E(\mathbf{X}_0 + \boldsymbol{\delta})
\approx \mathbf{F}(\mathbf{X}_0) - \mathbf{H} \boldsymbol{\delta}.
\label{eq:force_taylor}
\end{equation}

Eqs.~\eqref{eq:pes_taylor} and \eqref{eq:force_taylor} therefore apply at arbitrary reference configurations, including those along reaction paths. For small displacements, higher-order anharmonic contributions are negligible, so the second-order expansion closely reproduces the reference electronic structure energies
and forces.
Consequently, a single Hessian evaluated at a reference geometry provides a physically consistent local surrogate for both energy and forces, since Eq.~\eqref{eq:force_taylor} is the exact gradient of Eq.~\eqref{eq:pes_taylor}. We use this as the basis for our label synthesis in data augmentation.

\subsection{Data augmentation in machine learning}


Since energy and forces are continuous functions of the atomic coordinates, almost any perturbation changes the labels, unlike the label-preserving augmentations common in computer vision~\citep{shorten2019survey}. 
The exceptions are rigid translations and rotations, which leave all interatomic distances unchanged. Such augmentation benefits non-equivariant architectures~\citep{hu2021forcenet}, but is redundant for equivariant models, which satisfy these symmetries by construction.

The challenge in expanding molecular datasets therefore lies not in generating new configurations, but in obtaining accurate labels for them efficiently. ANI-1 introduced normal mode sampling to generate off-equilibrium structures by perturbing equilibrium geometries along their normal modes~\citep{smith2017ani}.  Although this approach exploits local curvature to sample structures, it does not exploit it to label them. Every generated structure still requires a separate quantum chemical evaluation, so the cost scales with the size of the augmented dataset.

Denoising pre-training instead avoids labeling cost by learning from unlabeled perturbations~\citep{zaidi2023pretraining}. Injecting Gaussian noise into equilibrium configurations and training a model to denoise them yields an objective equivalent to learning a force field under an isotropic Gaussian approximation of the Boltzmann distribution~\citep{zaidi2023pretraining}. Because isotropic noise assumes identical curvature along all spatial directions, subsequent works introduced coordinate-dependent noise schedules to mitigate this physical mismatch~\citep{feng2023fractional, ni2024sliced}. Denoising objectives have also been extended to off-equilibrium structures by conditioning the model on reference forces~\citep{liao2024generalizing}. These methods incorporate curvature or force cues into the training signal, but noise prediction remains an auxiliary self-supervised objective. The perturbed configurations themselves never acquire energy or force labels. In both cases, curvature informs where structures are sampled but not how they are labeled, so the labels are either expensive to obtain or absent altogether.

\begin{table}[htbp]
\centering
\caption{Datasets used in this work. Energies are in eV, forces in
  eV\,\AA$^{-1}$, and Hessians in eV\,\AA$^{-2}$.}
\label{tab:data_statistics}
\renewcommand{\arraystretch}{1.3}
\small
\setlength{\tabcolsep}{5pt}
\begin{tabular}{l l l l}
\toprule
 & \textbf{HORM}~\citep{cui2025large} & \textbf{HessianQM9}~\citep{williams2025hessian} & \textbf{MD17}~\citep{fu2023forces} \\
\midrule
Role            & \makecell[l]{Train + eval\\(in-domain and OOD)} & Train + in-domain eval & Zero-shot OOD eval \\
Subsets         & 3 (splits) & 4 (solvents) & 8 molecules (4 for MD) \\
Structures      & 1{,}836{,}206 & 41{,}645 $\times$ 4 & 8 trajectories \\
Regime          & Off-equilibrium & Equilibrium & MD trajectory \\
Hessian         & Yes & Yes & No \\
Level of theory & $\omega$B97X/6-31G(d) & $\omega$B97X/6-31G(d) & PBE+vdW-TS \\
\bottomrule
\end{tabular}
\end{table}
\section{Methods}

\paragraph{Hessian-based Augmentation Schemes}

Given a reference configuration $\mathbf{X}$ and its perturbed structure $\mathbf{X} + \boldsymbol{\delta} \mathbf{X}$, the corresponding changes in energy $\Delta E$ and forces $\Delta \mathbf{F}$ are approximated via a second-order Taylor expansion as

\begin{equation}
\Delta E = -\mathbf{F}^{\top}\delta \mathbf{X} + \frac{1}{2}\,\delta \mathbf{X}^{\top}\mathbf{H}\,\delta \mathbf{X},
\qquad
\Delta \mathbf{F} = -\mathbf{H}\,\delta \mathbf{X},
\label{eq:aug_taylor}
\end{equation}

where $\mathbf{F} = -\nabla{E(\mathbf{X})}$ and $\mathbf{H} = \nabla^2{E(\mathbf{X})}$ are the force vector and Hessian, respectively.


We propose two augmentation schemes, \textbf{UniAug} and \textbf{ModeAug}, which differ in the perturbation distribution from which $\boldsymbol{\delta}\mathbf{X}$ is sampled.

\paragraph{Uniform Cartesian augmentation (UniAug)}
UniAug samples atomic displacements directly in Cartesian space from an isotropic Gaussian distribution. For a given molecule with $N$ atoms, displacements are applied to a randomly selected subset $S$ of atoms, where $|S| = \lceil f \cdot N \rceil$ for a fraction $f \in (0, 1]$:
\begin{equation}
\boldsymbol{\delta} \mathbf{X}_i \sim \mathcal{N}(\mathbf{0},\,\sigma^2 \mathbf{I}_3) \quad (i \in S),
\qquad
\boldsymbol{\delta} \mathbf{X}_i = \mathbf{0} \quad (i \notin S),
\label{eq:uniaug}
\end{equation}
where the displacements are sampled independently for each atom and $\sigma$ controls the magnitude.

\paragraph{Normal mode augmentation (ModeAug)}

ModeAug samples displacements along its vibrational eigenvectors using the normal mode of molecule. Specifically, the mass-weighted Hessian $\mathbf{M}^{-1/2} \mathbf{H} \mathbf{M}^{-1/2}$ is diagonalized to obtain eigenvalues $\{\lambda_k\}$ and eigenvectors $\{\mathbf{v}_k\}$, and the translational and rotational modes are excluded, leaving $n$ vibrational modes indexed by $\mathcal{K}$. In practice, excluded modes are identified by $|\lambda_k| < 10^{-2}$. To ensure energy equipartition across modes, the amplitude of each mode is scaled by its inverse frequency $|\lambda_k|^{-1/2}$. The weights are then $L_2$-normalized over $\mathcal{K}$ and multiplied by $\sqrt{n}$. The resulting displacement $\boldsymbol{\delta}\mathbf{X}$ is constructed as below:
\begin{equation}
\boldsymbol{\delta}\mathbf{X} = \mathbf{M}^{-1/2} \sum_{k \in \mathcal{K}} c_k\,\mathbf{v}_k,
\quad
c_k = \sigma\,\varepsilon_k\,w_k,
\quad
w_k = \sqrt{n}\,\frac{|\lambda_k|^{-1/2}}{\big(\sum_{j\in\mathcal{K}}|\lambda_j|^{-1}\big)^{1/2}},
\quad
\varepsilon_k \sim \mathcal{N}(0,1),
\label{eq:modeaug}
\end{equation}
where $\mathbf{M}$ is the diagonal mass matrix and $\sigma$ controls the overall displacement scale. For both UniAug and ModeAug, the synthesized energy and force labels $(\Delta E, \Delta \mathbf{F})$ are computed via Eq.~\eqref{eq:aug_taylor}. An overview of these augmentation schemes is illustrated in Figure~\ref{fig:aug_overview_schemes}, with the resulting distributions of spatial displacements and potential energy shifts presented in Figure~\ref{fig:aug_overview_rmse}.

\paragraph{Comparison of the two schemes}
The two schemes offer complementary ways of exploring the configuration space. UniAug samples displacements independently of the local potential surface, resulting in augmented structures that are distributed isotropically in Cartesian space. In contrast, ModeAug assigns larger displacements to low-frequency modes and smaller ones to stiff, high-frequency modes, thereby distributing the perturbation energy uniformly across vibrational modes.


\paragraph{Role of $\sigma$.}
A larger augmentation magnitude explores broader regions of the potential energy surface, but the neglected higher order terms of the Taylor expansion grow accordingly. The two schemes interpret $\sigma$ differently: In UniAug, it denotes the standard deviation of Cartesian displacement components, whereas in ModeAug, it sets the scale of the mass weighted normal mode coefficients. Nevertheless, $\sigma$ controls the displacement magnitude in both schemes. We therefore tune $\sigma$ for each dataset and scheme by inspecting the resulting distributions of energy and force, and fix it rather than using a common value.

\subsection{Datasets and training}
We validated the effectiveness of Hessian-based data augmentation on three molecular benchmarks which cover distinct regions of the potential energy surface (Table~\ref{tab:data_statistics}). \textbf{HORM}~\citep{cui2025large} samples non-equilibrium and transition-state configurations along reactive pathways, \textbf{HessianQM9}~\citep{williams2025hessian} covers near-equilibrium geometries where net forces approach zero and local curvature governs harmonic vibrations, and \textbf{MD17}~\citep{chmiela2017machine} provides \textit{ab initio} molecular dynamics trajectories sampling thermal fluctuations. We use the MDsim reformulation~\citep{fu2023forces}, which redefines the data splits and evaluation protocol for assessing simulation stability.

Throughout this work, E-F denotes the standard setting in which models are trained with energy and force losses only, and E-F-H denotes the setting that additionally includes an explicit Hessian loss term. Because our augmentation scheme modifies only the training data and not the loss function, all our models are trained in the E-F setting, with the exception of E-F-H (RowSample), which is introduced solely as a baseline in the experiments that use only 10\% of the training data. Details of the datasets, evaluation protocols, and implementation are provided in the Supporting Information.

\paragraph{HORM}
HORM provides DFT-level energies, forces, and Hessians for non-equilibrium structures along reaction pathways, built from Transition-1x (\texttt{ts1x})~\citep{schreiner2022transition1x} and RGD1~\citep{zhao2023comprehensive}. Following the official protocol, models are trained on \texttt{ts1x-train} under both direct-force and autograd-force protocols and evaluated in domain on \texttt{ts1x-val}. RGD1 contains larger molecules than \texttt{ts1x} and forms a nearly disjoint distribution in chemical space, providing a chemical domain unseen during training. For zero-shot OOD evaluation, the trained models are applied to RGD1 without fine-tuning.

\paragraph{HessianQM9}
HessianQM9 provides numerical Hessians for equilibrium geometries of QM9 molecules, computed in vacuum and in three solvents treated with an implicit solvation model. The ground-truth Hessians are obtained at the same level of theory as HORM ($\omega$B97X/6-31G*). Models are trained separately for each of the four environments.

\paragraph{MDsim}
Following the MDsim protocol~\citep{fu2023forces}, single-point force accuracy is reported for all eight molecules, and 300 ps NVT rollout simulations are conducted on four of them to assess stability. In this benchmark, stability is quantified by $\#fin$, the number of trajectories completing the full simulation without any bond breaking, and structural fidelity by
$\mathrm{MAE}_{h(r)}$, the deviation between predicted and reference pair distance distributions. Because the MD17 dataset does not include reference Hessians, models trained on HORM (\texttt{ts1x-train}) are evaluated in a zero-shot cross-dataset transfer setting without fine-tuning. 

\subsection{Validation and feature changes of augmented structures}

In addition to measuring prediction accuracy on benchmarks, we examined the structural validity of the augmented molecules and analyzed the characteristics of energy and force changes using several methods. First, we compared the structural distribution of the perturbations and the resulting energy shifts induced by each augmentation mode. Second, we validated the energy changes from the Taylor approximation against DFT energies computed at the same level of theory as the original data, and related their agreement to the degree of structural deformation.
Third, we assessed the quality of the augmented data and labels by comparing the inferred forces from a model trained on perturbed structures against those directly calculated via DFT at the same level of theory as the HORM dataset. Lastly, we compared the degree of structural deformation by perturbation on a per-bond basis.

\paragraph{Structural distribution (t-SNE)}
To observe how the two augmentation schemes differ in configuration space, we visualize the distribution of atomic displacements induced by each perturbation with t-SNE. To inspect distribution, we randomly selected 30 molecules from each of the three HORM subsets (\texttt{ts1x-train}, \texttt{ts1x-val}, RGD1) and apply UniAug and ModeAug at $\sigma \in \{0.05, 0.1\}$\,\AA\  with three random seeds per molecule, and project the resulting displacements $\boldsymbol{\delta}\mathbf{X}$ after representing each as a fixed-dimensional descriptor vector. The procedure for constructing descriptor vectors of equal dimension from molecules with different numbers of atoms is described in the Supporting Information.

Additionally, we inferred the energies of both the sampled original molecules and their perturbed structures using a model trained exclusively on original data to compare the distribution of inferred energy variations induced by perturbations across different augmentation methods and scales.

\paragraph{Validation of Energies from Taylor Expansion Against DFT}

To examine whether the energies obtained from the Taylor approximation are accurate enough to serve as training data, we compared them against energies computed with DFT. We first randomly sampled 30 molecules from each subset of the HORM dataset. For each molecule, each augmentation scheme generated three structures using different random seeds, and this was repeated at three noise levels ($\sigma = 0.01$, $0.03$, and $0.05$). This yielded 90 structures per $\sigma$ and scheme in each subset. We then computed the energies of these augmented structures at the $\omega$B97X/6-31G* level, matching the level of theory with HORM, and compared them with the corresponding approximation. Agreement between the DFT energy change ($\Delta E_{\mathrm{DFT}}$) and the approximated energy change ($\Delta E$) was quantified with $R^2$. To relate this agreement to the degree of structural deformation, we also compared the RMSD from the original structure and the RMSD among structures generated from the same original structure (div).

\paragraph{Force shift ($\Delta \mathbf{F}$)}
We examine the magnitude of the force shift induced by each augmentation scheme and $\sigma$ against DFT references, and compare how well models trained on augmented data recover this response. The analysis uses the same sets of original and augmented molecules used in the t-SNE analysis for HORM. We performed the same analysis on both nonequilibrium structures from HORM and equilibrium structures from HessianQM9, sampling 30 molecules per split from each. The displacement scale was set to $\sigma \in \{0.01, 0.03, 0.05\}$\,\AA\ for HORM and $\sigma \in \{0.002, 0.01, 0.02\}$\,\AA\ for HessianQM9. We further compare the predicted $\lvert\mathbf{F}\rvert$ and $\Delta F$ distributions from three model checkpoints, trained without augmentation, with UniAug, and with ModeAug, against the reference DFT distribution. This serves as an accuracy check for how well each model recovers the true force response.

The force change $\Delta \mathbf{F}$ and force shift magnitude $\Delta F$ are defined as follows:
\begin{equation}
\Delta \mathbf{F} = \mathbf{F}(\mathbf{X} + \boldsymbol{\delta}\mathbf{X}) - \mathbf{F}(\mathbf{X}), \quad
\Delta F = \left(N^{-1}\sum_{i=1}^{N}\lVert \Delta\mathbf{F}_{i}\rVert^{2}\right)^{1/2}
\end{equation}

First, in original molecular structures, we predicted the distribution of absolute force values using four methods with three model checkpoints: (i)~model trained without augmented data (no aug), (ii)~model trained with UniAug, (iii)~model trained with ModeAug, together with (iv)~reference DFT at the same level of theory as the dataset---and compared the resulting distributions. This comparison of $\|\mathbf{F}\|$ serves not only as an accuracy metric for similarity to the DFT reference, but also as a verification of whether each model can achieve generalizability to sufficiently cover force distributions across various scales without bias.

Second, across various sets of augmented molecular structures, we grouped three approaches for each set: (i)~no aug, (ii)~UniAug or ModeAug, and (iii)~DFT calculations. We then plotted the force shift distributions by group according to each subset, augmentation mode, and $\sigma$ value. Through these plots, one can directly compare how the actual molecular forces change according to each augmentation mode and scale (reference DFT), and how the predictive capability for these force shifts varies depending on the inclusion of augmented data during training.

\paragraph{Bond deformation analysis}
This analysis examines whether each augmentation mode preserves the original covalent bonds and induces physically plausible fluctuations. To this end, we measured the change in bond lengths within each molecule before and after augmentation. Using the same set of molecules as in the above analysis, we collected all interatomic distances satisfying covalent bonding conditions and computed the relative deformation ratio $\Delta r / r = (r_{\mathrm{disp}} - r_{\mathrm{orig}}) / r_{\mathrm{orig}}$ before and after perturbation for histogram comparison. The distance-based bond criterion and its validity are detailed in the Supporting Information.

\subsection{Curvature learned by models}
\paragraph{Curvature Analysis Protocol}
To examine whether augmented data faithfully recover the true Hessian beyond improving energy and force predictions, we assess the quality of $\mathbf{H}_{\mathrm{pred}}$ obtained by differentiating predicted forces from six models trained on the HORM training set (no aug, UniAug, ModeAug, each under both direct force training and energy-conserved training). We report four metrics, three of which are in units of $\mathrm{eV}/\mathrm{\AA}^{2}$. Hessian MAE is the element-wise mean absolute error relative to reference values, and asymmetry error is $\frac{1}{9N^{2}}\sum_{k,l}\lvert\mathbf{H}_{kl} - \mathbf{H}_{lk}\rvert$~\citep{cui2025large}. 
Both metrics are evaluated on the raw Jacobian prior to symmetrization. For spectral metrics, the symmetric component $\frac{1}{2}(\mathbf{H}_{\mathrm{pred}} + \mathbf{H}_{\mathrm{pred}}^{\top})$ is preserved, leaving conservative models unchanged. Eigenvalue MAE compares the full $3N$ signed spectra paired by ascending indices without removing the translational and rotational modes, that is, the six smallest modes. In contrast, $n_{\mathrm{imag}}$ counts the number of negative eigenvalues of the mass weighted Hessian after excluding these six modes.

For evaluation, 2,000 molecules uniformly sampled at random with a fixed seed from \texttt{ts1x-val} and RGD1 were used. 
Comparisons were also made against E--F and E--F--H values reported in the original dataset paper~\citep{cui2025large}. The upstream EquiformerV2 implementation detaches the local rotation frame from the autograd graph, and the effect of this setting on the curvature metrics is described in the Supporting Information.

\paragraph{Qualitative case study}
In addition to the quantitative metrics regarding Hessian recovery, we examined individual molecular cases. Among the \texttt{ts1x-val} molecules used in the curvature analysis, we selected and plotted two molecules for each category where the ModeAug-trained model showed \{low, similar, high\} eigenvalue prediction accuracy relative to the no aug model, specifically within the stiff mode regime ($400$--$3800\,\mathrm{cm}^{-1}$) corresponding to stretching vibrations derived from each molecule's Hessian (detailed procedures are provided in the Supporting Information).

\subsection{Computational cost and efficiency}
We analyze the computational cost and efficiency of the proposed data augmentation scheme. Specifically, we compare it against two Hessian-supervised pathways that evaluate the model's own force Jacobian during training (full Hessian supervision vs.~stochastic row subset supervision)~\citep{cui2025large}. All comparison criteria---curvature source, inclusion in the autograd graph, number of backward passes, and retained-graph memory---represent model-agnostic metrics independent of specific model architectures.

\section{Results}

This section is organized as follows. The first part analyzes the augmented data itself, comparing how the two schemes differ in the structures they generate and how faithfully the resulting labels track reference DFT values. Downstream performance is then reported on HORM, HessianQM9, and MD17, followed by an analysis of PES curvature and computational cost.

\subsection{Validity and diversity of augmented data}

This subsection characterizes the augmented data itself, assessing whether the two schemes produce physically valid labels and whether they explore distinct regions of the PES. Reference DFT calculations serve as the standard against which the model-predicted quantities are assessed.

\paragraph{Displacement diversity by augmentation mode}

\begin{figure}[htbp]
  \centering
  \includegraphics[width=\linewidth]{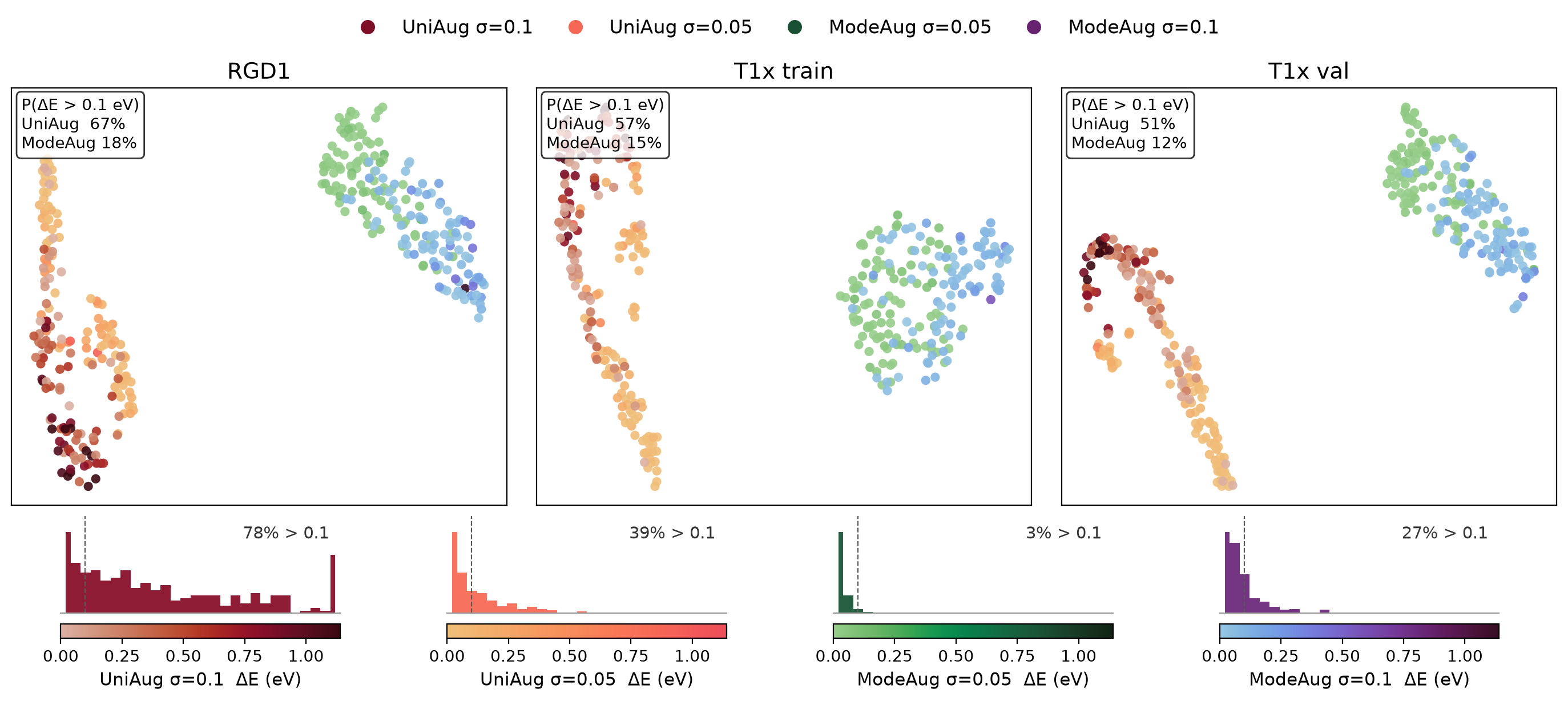}
    \caption{t-SNE projections of augmentation displacement fields across data subsets. Hue distinguishes the four scheme\,$\times\,\sigma$ combinations, while color intensity encodes the predicted $\Delta E$ (lighter indicates lower energy shift). Inset boxes report $P(\Delta E > 0.1\,\mathrm{eV})$ for each scheme, and bottom histograms display the corresponding $\Delta E$ distributions.}

  \label{fig:tsne}
\end{figure}

Examining the projected regions of vectors representing atomic displacements for the same molecule sets according to augmentation strategy and scale in Figure~\ref{fig:tsne} shows that the two augmentation strategies are distinctly separated in the projected manifolds, with further distributional variations occurring across different scales. This distinction is also prominent in the results where $\Delta E$ for each panel is color-encoded. On \texttt{ts1x-train}, 78\% of UniAug samples at $\sigma = 0.1$\,\AA{} and 39\% at $\sigma = 0.05$\,\AA{} exceed $\Delta E > 0.1$\,eV, whereas ModeAug reaches only 27\% and 3\% at the same scales. UniAug therefore displaces atoms toward higher-energy regions than ModeAug, and increasing $\sigma$ shifts both schemes in the same direction.


\paragraph{Taylor Approximation Accuracy Across Augmentation Strategies}

First, comparing across values of $\sigma$, both UniAug and ModeAug showed a tendency for the coefficient of determination $(R^2)$ to decrease slightly as $\sigma$ increased. This tendency can be attributed to the fact that both the RMSD and div increase with $\sigma$, which enlarges the structural difference from the original and increases the error of the Taylor approximation.

Next, comparing the two augmentation schemes, ModeAug achieved a higher $R^2$ than UniAug at every $\sigma$ level and in every subset. At $\sigma = 0.01$, both schemes gave $R^2 > 0.999$ with almost no difference. As $\sigma$ increased, however, the $R^2$ of UniAug dropped to 0.970 to 0.989 at $\sigma = 0.05$, whereas that of ModeAug remained at 0.997 to 0.998. The standard deviation (SD) of the actual $\Delta E_{\mathrm{DFT}}$ was also consistently larger for UniAug than for ModeAug (0.13 to 0.19 eV versus 0.04 to 0.05 eV at $\sigma = 0.05$), which shows that, as intended by design, the range of energy change for ModeAug is not as large as that for UniAug. This pattern persisted even though ModeAug had a larger RMSD (about 0.035 to 0.039~\AA\ versus 0.022 to 0.023~\AA\ at $\sigma = 0.05$), suggesting that the accuracy of the Taylor expansion depends more on the direction in which the deformation is applied than on the magnitude of the deformation itself. Consequently, ModeAug maintained a higher Taylor approximation accuracy while achieving greater structural diversity than UniAug.

\begin{figure}[htbp]
  \centering
  \includegraphics[width=0.98\linewidth]{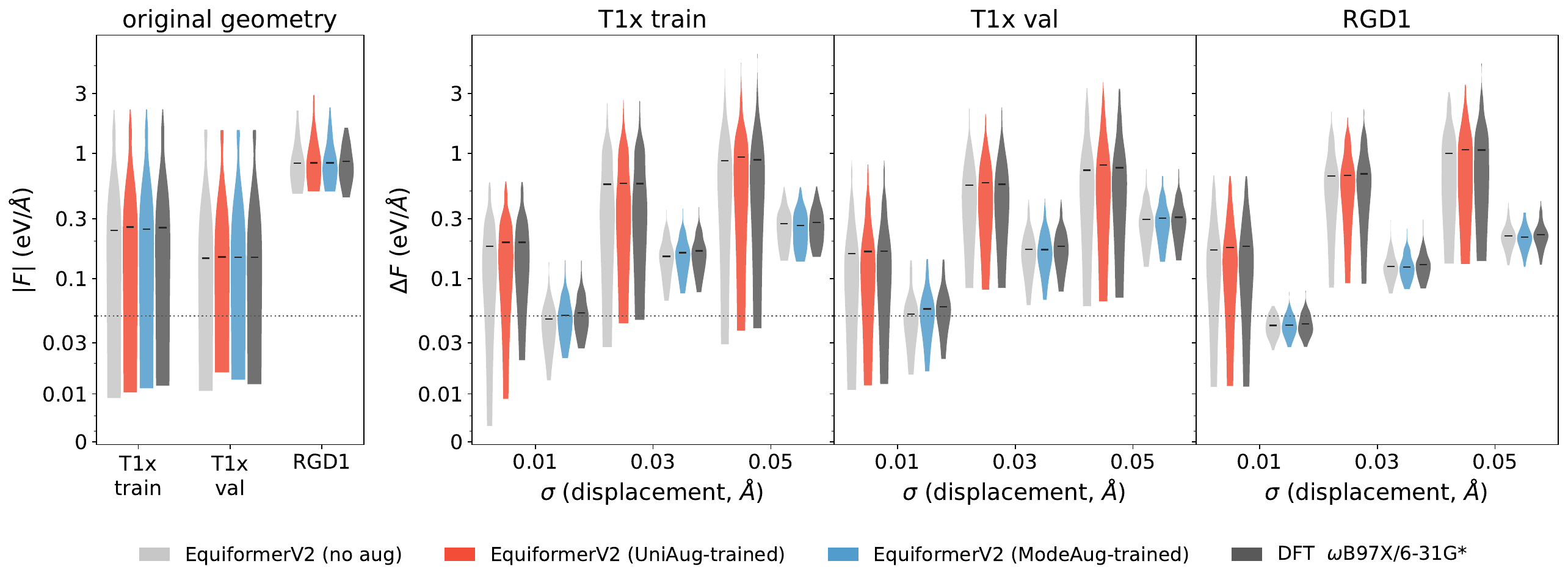}
  \caption{Distribution of force changes $\Delta \mathbf{F}$ under displacement on the three HORM subsets. \textbf{Left}: Force magnitude $|F|$ at the original geometry, comparing the dataset label with the No-aug model prediction. \textbf{Right}: $\Delta \mathbf{F} = \mathbf{F}(\mathbf{X} + \boldsymbol{\delta \mathbf{X}}) - \mathbf{F}(\mathbf{X})$ at $\sigma \in \{0.01, 0.03, 0.05\}$\,\AA, grouped by displacement type. Within each group, the three violins represent the No-aug model (left), the augmented model (middle), and DFT (right).}
  \label{fig:violin_plot_horm}
\end{figure}

\paragraph{Force shift distribution recovery (HORM)}
Figure~\ref{fig:violin_plot_horm} presents the $\Delta F$ distributions across different subsets of HORM. 

\textbf{Left.} The left panel represents evaluations on original molecular structures before augmentation. In the ts1x-train set, model checkpoints trained on ModeAug, UniAug, and No aug data cover the $\lvert\mathbf{F}\rvert$ distribution of dataset labels in that order. Comparing inference distributions on \texttt{ts1x-val} and RGD1, UniAug shows the largest discrepancy from the reference DFT distribution, whereas no aug and ModeAug recover it more closely. Overall, the predicted distributions from the three models remain closely aligned on the original dataset.

\textbf{Right.} The results in the right panel evaluated on augmented data demonstrate clear differences between UniAug and ModeAug across all three subsets. Comparing the reference DFT values first, UniAug clearly induces significantly larger force shifts than ModeAug across all subsets and $\sigma$ values. When comparing how well the no aug model and the models trained on each augmentation mode cover this DFT shift distribution, models trained on augmentation modes recover it better than the no aug model in almost all cases.

These results imply the following: (i)~models trained with augmented data achieve more accurate force predictions on both original and augmented data compared to models trained without augmentation, (ii)~UniAug induces noticeably larger force variations than ModeAug, which is an expected outcome when considering the vibrational characteristics of molecules, and (iii)~as the perturbation scale ($\sigma$) increases, the magnitude of force shifts increases for both augmentation methods, which can be attributed to force variations growing as structural deformations increase.

\begin{figure}[htbp]
  \centering
  \includegraphics[width=0.98\linewidth]{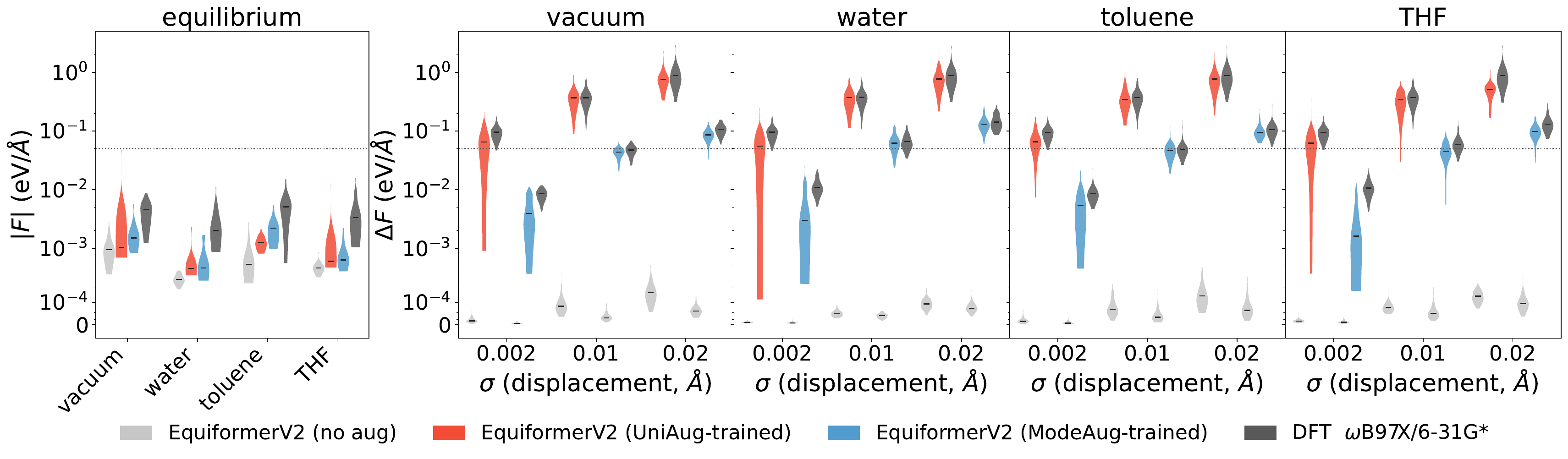}
    \caption{Distribution of force changes $\Delta \mathbf{F}$ under displacement on the four HessianQM9 environments at $\sigma \in \{0.002, 0.01, 0.02\}$\,\AA. Conventions are the same as those in Figure~\ref{fig:violin_plot_horm}, except that the $y$-axis scale is considerably smaller in HessianQM9. Note that dataset labels (left) and DFT (right) are both $\omega$B97X/6-31G* but from different codes (NWChem vs. GPU4PySCF), shown in one color for clarity.}
  \label{fig:violin_plot_hqm9}
\end{figure}

\paragraph{Equilibrium force shift distribution recovery (HessianQM9)}
Figure~\ref{fig:violin_plot_hqm9} shows the corresponding $\Delta F$ distribution for HessianQM9 in the same manner as Figure~\ref{fig:violin_plot_horm}. 

\textbf{Left.} On the unaugmented original dataset (left panel), although there are differences depending on the solvent type, models trained with UniAug and ModeAug consistently exhibit predicted distributions closer to the ground truth than the no aug model checkpoint. Unlike HORM, the HessianQM9 dataset consists of molecules in equilibrium structures, meaning most absolute force values are extremely small and close to zero. When compared, the medians of all model prediction distributions are lower than the reference. These models appear particularly vulnerable to zero-bias, failing to adequately distinguish molecules with relatively higher absolute force values. This tendency worsens as the reference median value itself becomes smaller; for the two solvents with the smallest values (water, THF), compared to vacuum and toluene (which have relatively larger absolute force median values), the predicted medians of all models fail to reach even the minimum values of the actual reference. This connects to the comparison of energy and force prediction performance between HORM and HessianQM9 discussed later.

\textbf{Right.} In the right panel evaluated on augmented data, characteristic differences across solvent types are minor and follow similar patterns in both reference DFT values and predicted values (Note that, unlike HORM---an off-equilibrium dataset where absolute force values are generally large---the $\sigma$ values were significantly reduced for this plot to reflect the fact that most force values are close to zero). Looking at the reference DFT as a benchmark, the actual magnitude of force shift increases with larger $\sigma$, and predictions from all three models appear to have learned this pattern to some extent. Examining the predicted distributions, underestimation in predictions becomes more severe with less augmentation. Considering this alongside the left panel results, for equilibrium sets, data augmentation itself helps the model break away from zero-bias, preventing prediction collapse while improving predictive power.

\begin{figure}[htbp]
  \centering
  \includegraphics[width=0.98\linewidth]{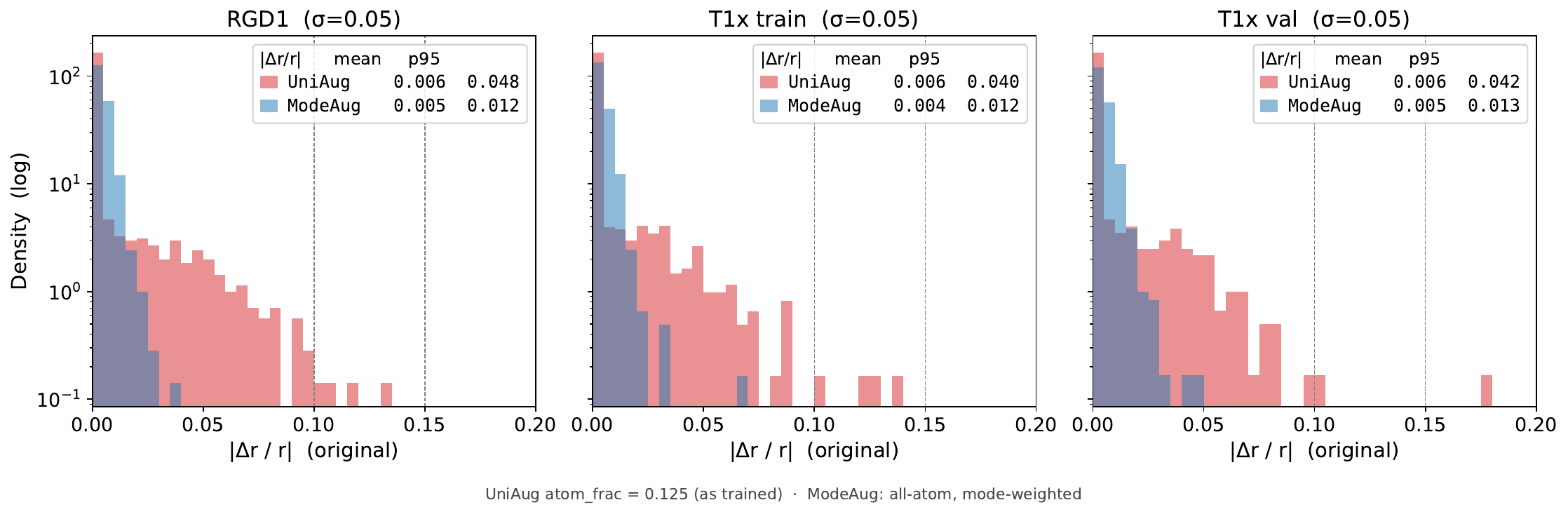}
  \caption{Distribution of relative bond deformation $|\Delta r/r|$ under perturbation ($\sigma = 0.05\,\mathrm{\AA}$) across HORM subsets, comparing UniAug and ModeAug. 
  }
  \label{fig:bond_deviation}
\end{figure}

\paragraph{Relative bond length deviation}
When comparing the range of mean relative bond deformation $\lvert \Delta r/r \rvert$ between the two augmentation strategies in Figure~\ref{fig:bond_deviation}, UniAug induces significantly broader and larger deformations (up to ${\sim}15\%$) compared to ModeAug (within ${\sim}5\%$) across all three HORM subsets, consistent with the larger energy and force shifts observed for UniAug in the preceding analyses. When comparing mean values, however, both UniAug and ModeAug average around $0.5\%$, indicating that for both strategies, the vast majority of augmentations cause minimal bond length deviations.

\subsection{Energy and force prediction}
\paragraph{HORM}

We investigated the impact of the augmentation ratio and displacement magnitude ($\sigma$) under direct force training. Increasing the ratio of original to augmented data from $1{:}1$ to $1{:}3$ offered no clear benefit in domain (\texttt{ts1x-val}). Out-of-domain, however, the force MAE for ModeAug dropped to $0.1017~\text{eV/\AA}$, which was the lowest among all direct force settings, suggesting that a higher proportion of augmented data can enhance generalization to unseen structures. Nonetheless, this benefit was confined to force prediction and provided little advantage for energy prediction.

Increasing $\sigma$ for ModeAug from $0.05$ to $0.10~\text{\AA}$ degraded both energy and force metrics in domain ($0.0109~\text{eV/\AA}$) and out-of-domain ($0.1041~\text{eV/\AA}$) alike. This supports our physical intuition that a larger displacement scale broadens the $\Delta E$ distribution beyond the local regime where the second order Taylor expansion remains valid. We note that the optimal scale of $0.05~\text{\AA}$ should be interpreted as specific to the HORM dataset, as the appropriate threshold may vary depending on the dataset.

Finally, a mixed augmentation strategy with an original to UniAug to ModeAug ratio of $2{:}1{:}1$ was evaluated. However, it failed to outperform either single method scheme across all metrics and is omitted from the results for brevity.

\begin{figure}[htbp]
  \centering
  \includegraphics[width=0.98\linewidth]{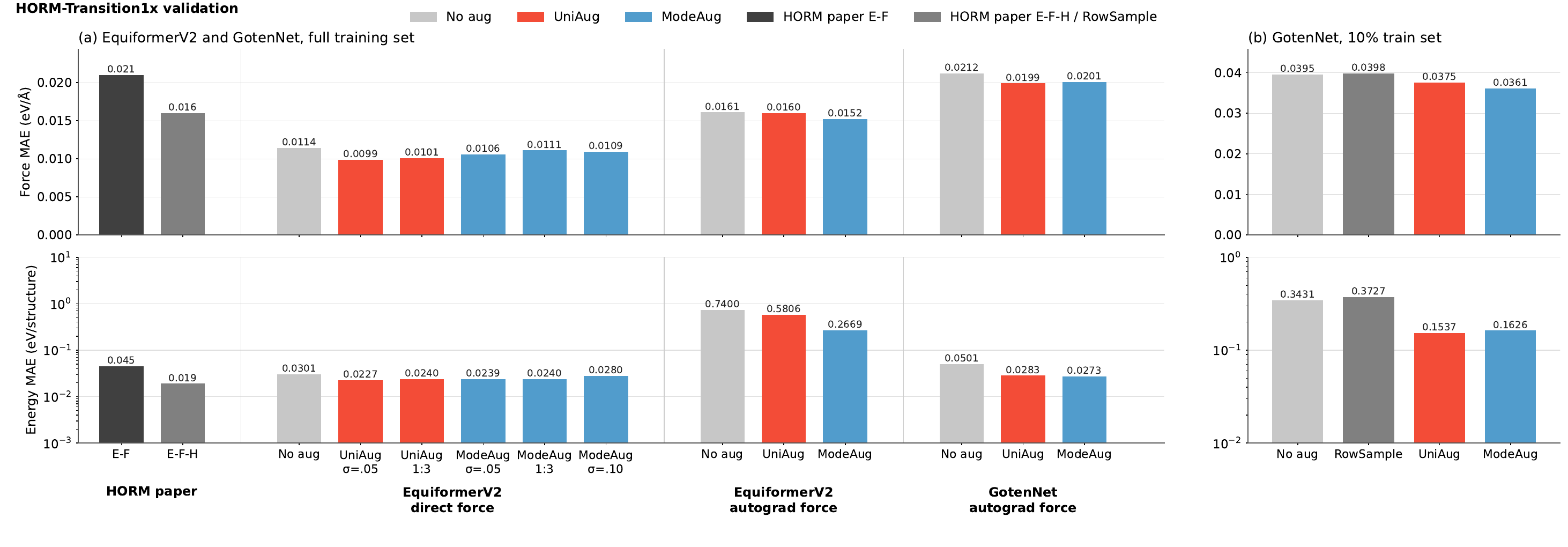}
  \caption{Force MAE (eV/\AA, top) and energy MAE (eV/structure, bottom, log scale) for each augmentation scheme on the HORM Transition1x validation set. (a) EquiformerV2 (direct force and autograd force) and GotenNet (autograd force) trained on the full training data. (b) GotenNet trained on 10\% of the training data. Gray denotes no augmentation, red UniAug, and blue ModeAug. Dark gray denotes the baseline from the HORM paper (E-F). Medium gray denotes the HORM baseline (E-F-H) in (a), and in (b) the E-F-H loss function of the HORM paper (RowSample) applied to GotenNet. $\sigma$ is the displacement magnitude, and 1:3 is the ratio of original to augmented samples in a batch. Unless otherwise noted, the ratio is 1:1 and $\sigma = 0.05$.}
  \label{fig:bench_ts1x}
\end{figure}

To examine how augmentation affects model performance, we compared the energy and force MAEs of each augmentation scheme on the HORM Transition1x validation set (Figure~\ref{fig:bench_ts1x}). For EquiformerV2 with direct force prediction trained on the full training data, augmentation reduced the energy MAE from 0.0301 eV to 0.0227--0.0280 eV and the force MAE from 0.0114 eV/\AA\ to 0.0099--0.0111 eV/\AA, with little difference between UniAug and ModeAug. For EquiformerV2 with autograd forces, in contrast, the effect of ModeAug was pronounced. The energy MAE was 0.7400 eV without augmentation, 0.5806 eV with UniAug (about 22\% lower), and 0.2669 eV with ModeAug (about 64\% lower). For GotenNet~\cite{aykent2025gotennet}, both schemes lowered the energy MAE from 0.0501 eV to 0.0283 eV (UniAug) and 0.0273 eV (ModeAug), nearly halving it, and the two schemes performed similarly. Thus, augmentation consistently reduced the energy error regardless of the model and the force computation method, whereas the improvement in force MAE was relatively small (about 1--6\% for autograd models).

With only 10\% of the training data, GotenNet showed clearer differences among the methods. UniAug and ModeAug lowered the energy MAE from 0.3431 eV to 0.1537 eV and 0.1626 eV (about 55\% and 53\%), and the force MAE from 0.0395 eV/\AA\ to 0.0375 eV/\AA\ and 0.0361 eV/\AA\ (about 5\% and 9\%), respectively. RowSample, in contrast, left the force MAE almost unchanged (0.0398 eV/\AA) and even increased the energy MAE to 0.3727 eV. These results show that when training data are limited, both UniAug and ModeAug offer a clearer benefit than learning the Hessian through supervision.

\begin{figure}[htbp]
  \centering
  \includegraphics[width=0.98\linewidth]{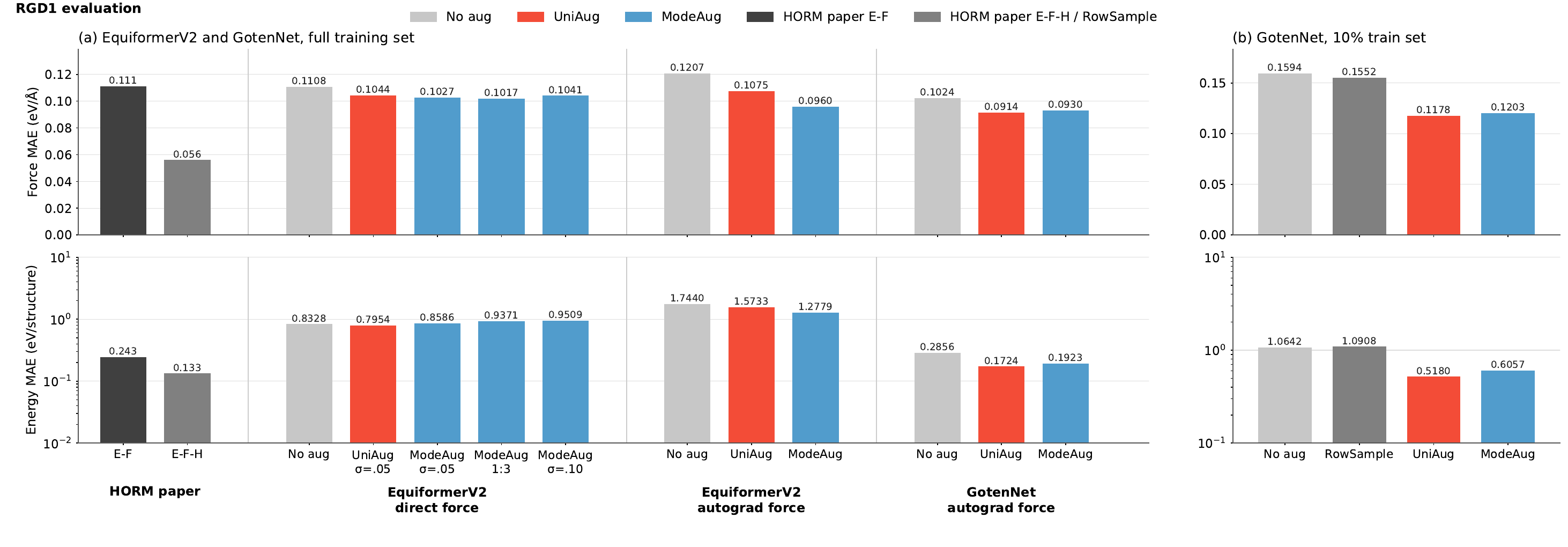}
  \caption{Force MAE (eV/\AA, top) and energy MAE (eV/structure, bottom, log scale) for each augmentation scheme on the RGD1 evaluation set. (a) EquiformerV2 (direct force and autograd force) and GotenNet (autograd force) trained on the full training data. (b) GotenNet trained on 10\% of the training data. The color coding is the same as in Figure~\ref{fig:bench_ts1x}}
  \label{fig:bench_rgd1}
\end{figure}

We also compared the energy and force MAEs of each augmentation scheme on the RGD1 evaluation set (Figure~\ref{fig:bench_rgd1}). For EquiformerV2 with direct force prediction trained on the full training data, augmentation reduced the force MAE from 0.1108 eV/\AA\ to 0.1017--0.1044 eV/\AA\ (about 6--8\%). The energy MAE decreased only with UniAug, from 0.8328 eV to 0.7954 eV (about 5\%), and instead increased to 0.8586--0.9509 eV with ModeAug. For EquiformerV2 with autograd forces, in contrast, the effect of ModeAug was clear. The force MAE was 0.1207 eV/\AA\ without augmentation, 0.1075 eV/\AA\ with UniAug (about 11\% lower), and 0.0960 eV/\AA\ with ModeAug (about 21\% lower).

With only 10\% of the training data, both UniAug and ModeAug improved performance consistently, as on the Transition1x validation set. UniAug and ModeAug lowered the energy MAE from 1.0642 eV to 0.5180 eV and 0.6057 eV (about 51\% and 43\%), and the force MAE from 0.1594 eV/\AA\ to 0.1178 eV/\AA\ and 0.1203 eV/\AA\ (about 26\% and 25\%), respectively. RowSample, in contrast, gave energy (1.0908 eV) and force (0.1552 eV/\AA) MAEs almost identical to those without augmentation. Even for this out of distribution set, the benefit of augmentation was clearer for the model trained with limited data, and the gains were generally larger when the model was trained with autograd forces.

\paragraph{HessianQM9}

Table~\ref{tab:hqm9_bestsigma} reports the validation performance across four solvent environments in HessianQM9. Unlike HORM, this dataset contains only equilibrium geometries. Consequently, the overall force magnitude $\langle\vert{}\mathbf{F}\vert{}\rangle$ is substantially smaller ($1.38$ to $1.93~\text{meV/\AA}$) than that of HORM, which included structures away from equilibrium. This value serves as a reference threshold to determine whether a model has learned a meaningful physical signal rather than collapsing to a trivial prediction where $\mathbf{F}=0$.

Notably, the value reported in the original paper is larger than ours. We attribute this discrepancy to their evaluation at the molecule level, unlike our per atom metric. Typically, the molecule-wise MAE is roughly $18.4$ times the atom-wise MAE, corresponding to the average number of atoms per molecule. Because we could not verify the exact calculation details of their metric, we suggest treating this comparison as a qualitative reference, unlike our direct comparison in HORM.

Among the twelve models trained across four solvents and three direct force training settings, only No aug and ModeAug yielded force prediction errors below $\langle\vert{}\mathbf{F}\vert{}\rangle$ in vacuum and toluene. In both solvents, ModeAug outperformed No aug, reducing the error by $3.6\%$ in vacuum and $6.2\%$ in toluene relative to $\langle\vert{}\mathbf{F}\vert{}\rangle$. UniAug failed to fall below the $\langle\vert{}\mathbf{F}\vert{}\rangle$ threshold in both solvents and performed worse than No aug.

In the cases of water and THF, however, no augmentation setting reached the baseline threshold. Although both augmentation schemes performed worse than No aug in these solvents, No aug itself exhibited an MAE comparable to $\langle\vert{}\mathbf{F}\vert{}\rangle$, indicating that all methods failed to learn a meaningful signal beyond the trivial solution. Although the exact cause of this failure cannot be directly established, we hypothesize that the lower $\langle\vert{}\mathbf{F}\vert{}\rangle$ values in water and THF resulted in sparser learning signals, preventing the models from distinguishing meaningful patterns from noise.

We report the optimal $\sigma$ values for each solvent determined by heuristic search. Unlike in vacuum and toluene, tuning $\sigma$ was not effective for improving predictive accuracy in water and THF.

\paragraph{MD17: Single-point force prediction}

As noted in the methodology section, this evaluation represents a zero shot test using the three model checkpoints trained with direct force on HORM without any further fine tuning. Table~\ref{tab:md17_stage1} shows that ModeAug achieved higher predictive accuracy than both No aug and UniAug across all eight molecules, whereas UniAug generally performed slightly worse than No aug. In the force cosine metric, which compares normalized predicted and reference forces, UniAug showed no significant improvement over No aug overall, while ModeAug attained the highest alignment score for six molecules. Considering that this is a zero shot evaluation, these results suggest that training with ModeAug on non equilibrium structures can modestly enhance generalization even to equilibrium molecules.

\paragraph{Nosé–Hoover MD simulation}
Table~\ref{tab:md17_nh} reports the results of $300~\text{ps}$ zero shot NVT simulations across four test molecules. We note that the same three model checkpoints evaluated in the single point force prediction above were used here for zero shot evaluation. Both augmentation schemes outperformed No aug, though they excelled along different evaluation axes.

In terms of trajectory stability, UniAug showed superior robustness, completing all 20 trajectories. ModeAug and No aug completed 16 and 14 trajectories, respectively. The failure patterns differed by model. No aug collapsed across all five runs for salicylic acid, resulting in a mean stability time of only $18.3~\text{ps}$. In contrast, the instability of ModeAug's instability was concentrated in ethanol, while it demonstrated stability comparable to UniAug

Regarding structural distribution accuracy, the performance ordering reversed, with ModeAug achieving the lowest $h(r)$ MAE across all four molecules. The mean $h(r)$ MAE values were $0.0050$ for ModeAug, $0.0058$ for UniAug, and $0.0085$ for No aug. The most substantial gain was observed for naphthalene, where ModeAug halved the error of No aug. Because No aug had no completed trajectories for salicylic acid, that molecule was excluded from its average, and comparisons with No aug are based on the remaining three molecules.

\subsection{PES curvature recovery}
Tables~\ref{tab:curvature_evaluation_ts1x} and \ref{tab:curvature_evaluation_rgd1} report potential energy surface (PES) curvature metrics for 2,000 molecules on the in-domain (\texttt{ts1x-val}) and out-of-domain (RGD1) test sets, respectively.

First, comparing the predictive performance of the models in Table~\ref{tab:curvature_evaluation_ts1x}, both direct force training and energy-conserved models showed improved predictive capability across all four metrics when trained with augmented data. Notably, models trained on ModeAug data exhibited superior performance metrics in most cases. When Hessian and eigenvalue predictions are compared with reported values from the original paper~\citep{cui2025large}, while performance does not reach the level of explicitly training on Hessians (E--F--H), but MAEs for both Hessians and eigenvalues are substantially reduced to roughly 20--30\% of their values compared to models trained without Hessians (E--F). However, because it cannot be confirmed whether all hyperparameters (such as loss term weights or learning rates) were identically configured, these results are best interpreted as indicating a substantial reduction in error rather than as exact numerical values.

Next, looking at Table~\ref{tab:curvature_evaluation_rgd1}, augmentation again consistently improves predictive capability, though the relative superiority of UniAug versus ModeAug varies across metrics. In particular, UniAug yielded better performance for Hessian and eigenvalue predictions in both direct force training and energy-conserved training, displaying a different trend from the \texttt{ts1x-val} results. Compared with the original paper's report, performance still falls short of E--F--H (which directly trains on Hessians), but achieves significantly lower error values than those reported for E--F.
In summary, both augmentation techniques intrinsically recover PES curvature-related information to a significant extent solely through Taylor-approximation-based augmentation, even when Hessian information is excluded during training. 

\begin{figure}[htbp]
    \centering

    \begin{subfigure}[b]{0.48\textwidth}
        \centering
        \includegraphics[width=\textwidth]{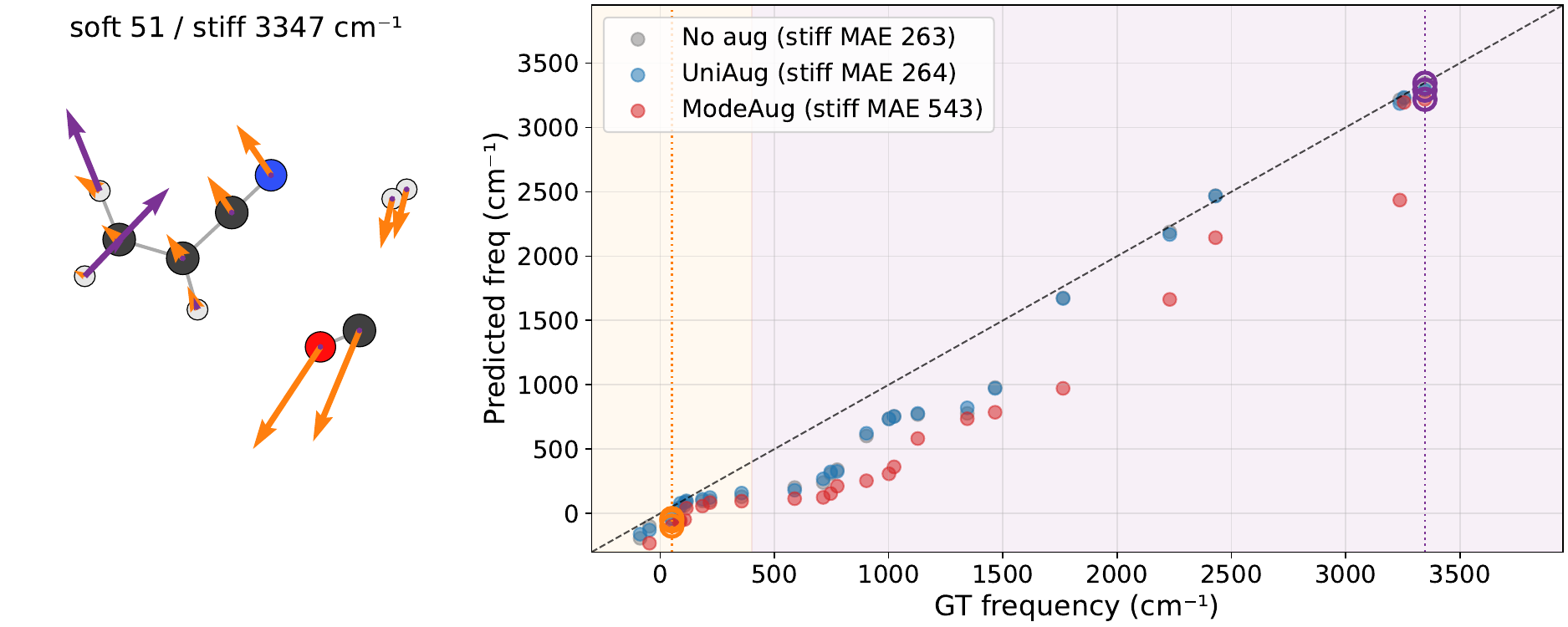}
        \caption{Fail (1)}
        \label{fig:curvature_case_study_fail1}
    \end{subfigure}
    \hfill
    \begin{subfigure}[b]{0.48\textwidth}
        \centering
        \includegraphics[width=\textwidth]{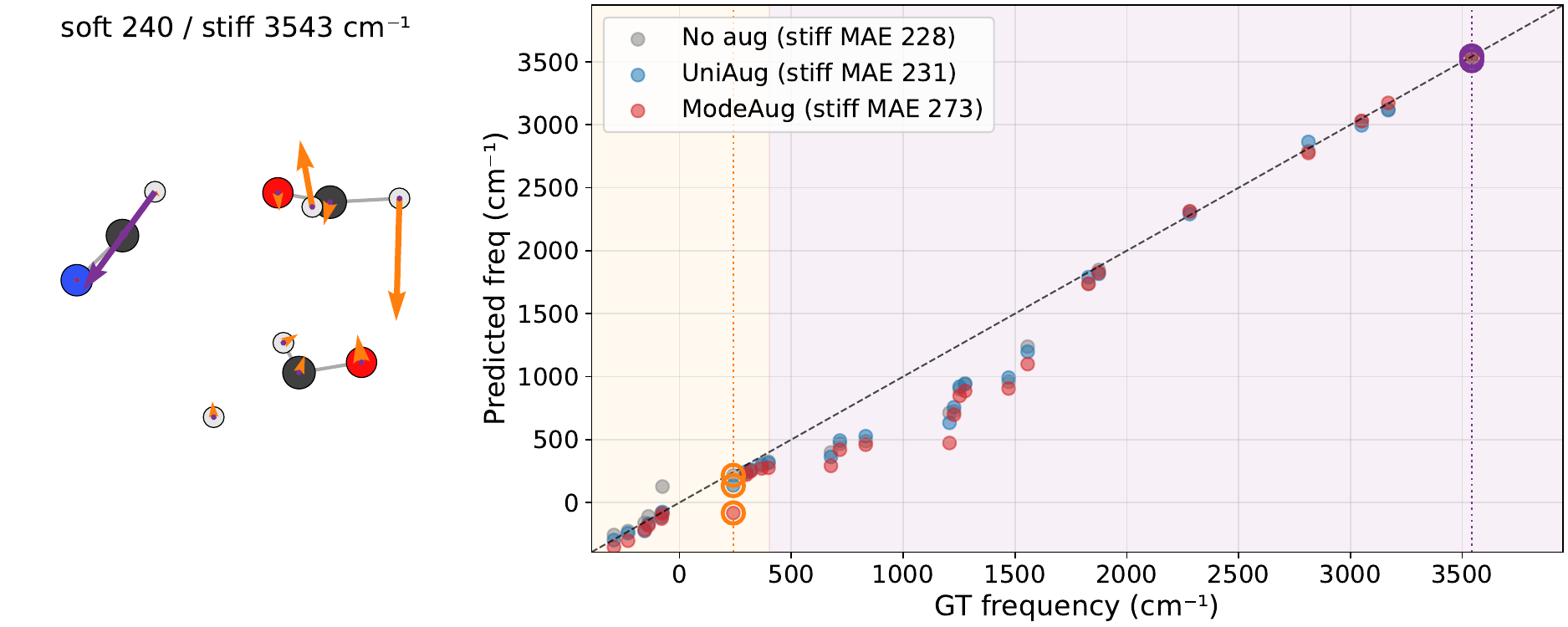}
        \caption{Fail (2)}
        \label{fig:curvature_case_study_fail2}
    \end{subfigure}

    \vspace{0.5em}

    \begin{subfigure}[b]{0.48\textwidth}
        \centering
        \includegraphics[width=\textwidth]{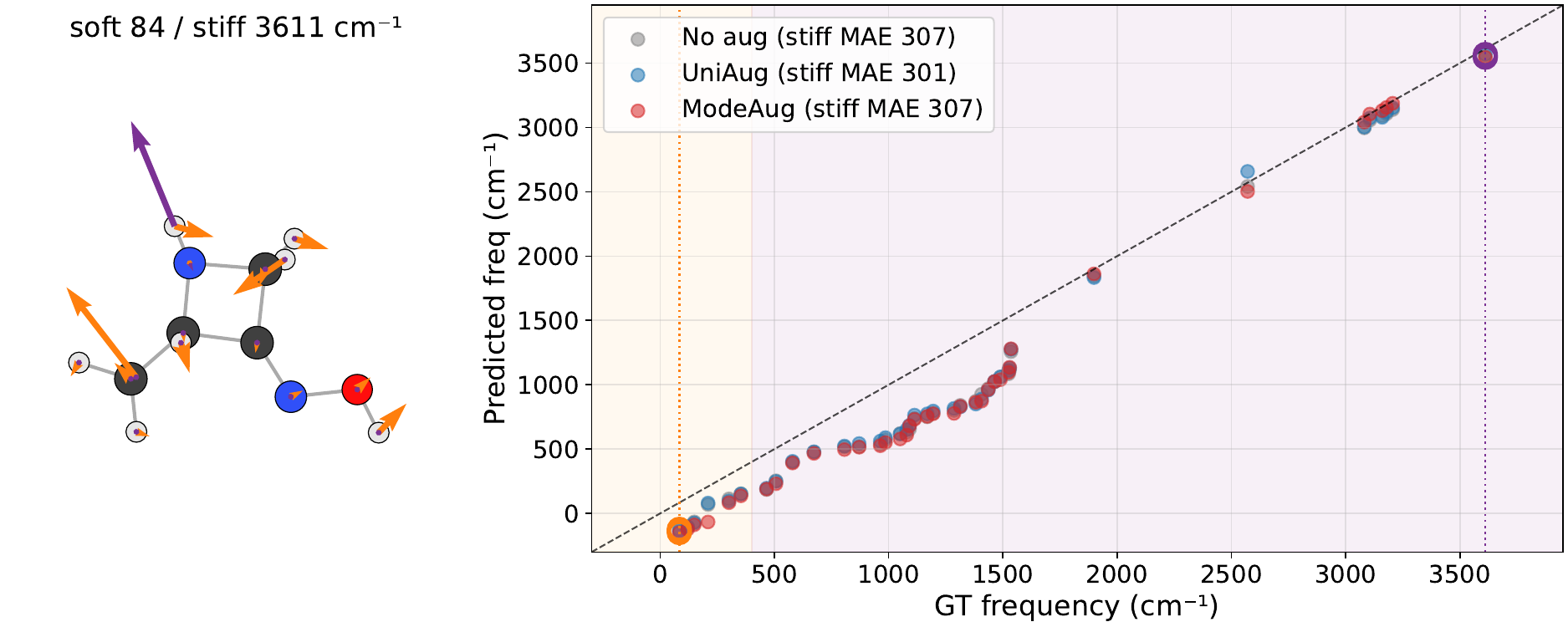}
        \caption{Normal (1)}
        \label{fig:curvature_case_study_normal1}
    \end{subfigure}
    \hfill
    \begin{subfigure}[b]{0.48\textwidth}
        \centering
        \includegraphics[width=\textwidth]{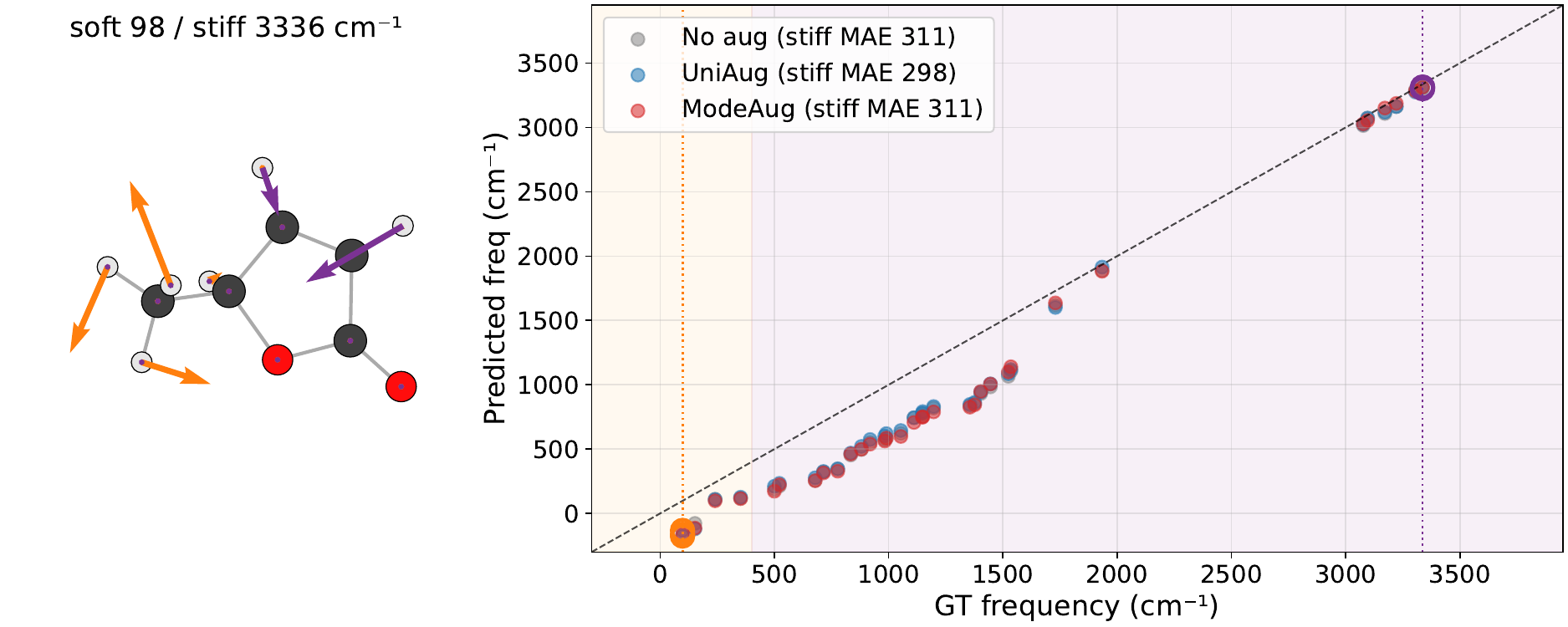}
        \caption{Normal (2)}
        \label{fig:curvature_case_study_normal2}
    \end{subfigure}

    \vspace{0.5em}

    \begin{subfigure}[b]{0.48\textwidth}
        \centering
        \includegraphics[width=\textwidth]{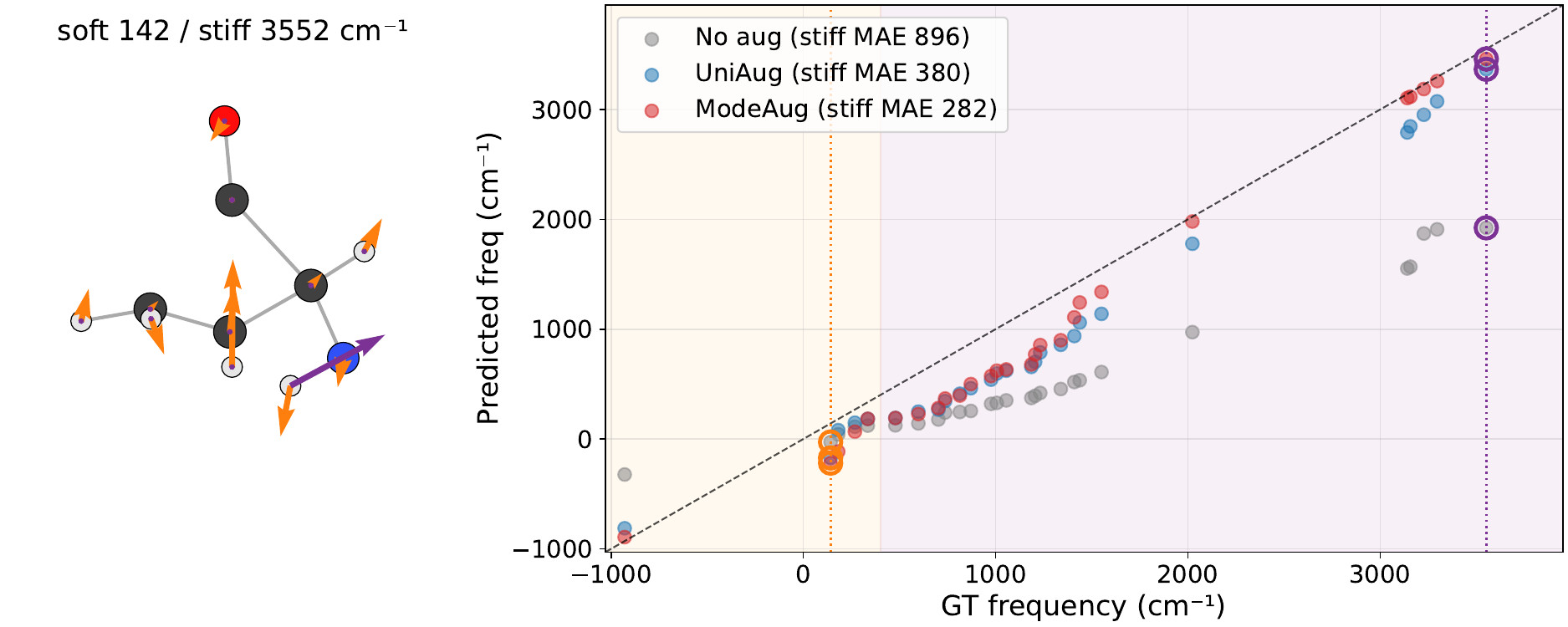}
        \caption{Success (1)}
        \label{fig:curvature_case_study_success1}
    \end{subfigure}
    \hfill
    \begin{subfigure}[b]{0.48\textwidth}
        \centering
        \includegraphics[width=\textwidth]{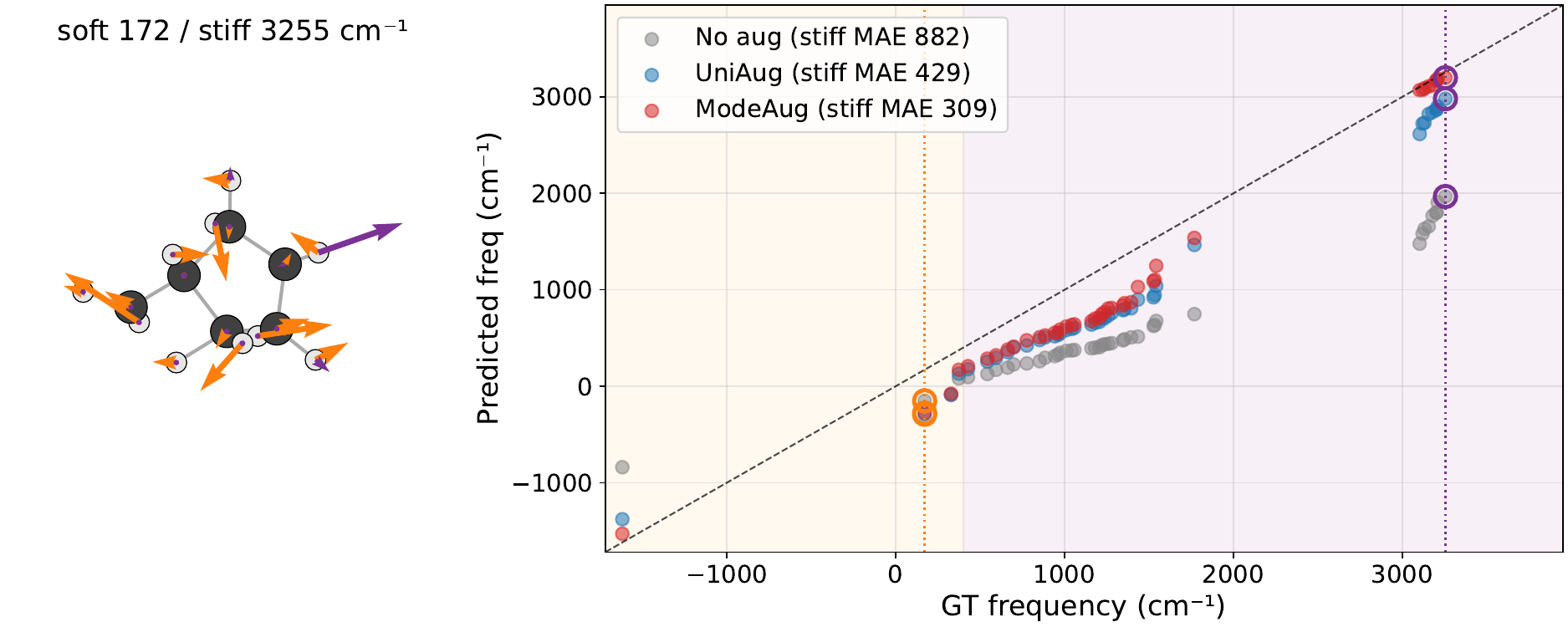}
        \caption{Success (2)}
        \label{fig:curvature_case_study_success2}
    \end{subfigure}

      \caption{Model-derived harmonic vibrational frequencies compared against reference DFT values for \textbf{(a, b)} failure, \textbf{(c, d)} normal, and \textbf{(e, f)} success cases. The dashed diagonal indicates $y=x$. Insets depict normal-mode displacement vectors. Frequencies are capped at $\leq 3800~\mathrm{cm^{-1}}$, with orange- and purple-shaded regions marking the soft ($<400~\mathrm{cm^{-1}}$) and stiff ($400$--$3800~\mathrm{cm^{-1}}$) bands, respectively, and the reported stiff MAE is computed over the stiff band $[400, 3800]~\mathrm{cm^{-1}}$.}
      
    \label{fig:curvature_case_study}

\end{figure}

\paragraph{Case study}
In the qualitative case study (Figure~\ref{fig:curvature_case_study}), the observed patterns across the two fail, normal, and success cases are as follows: In fail cases, we found that molecules with a higher number of low ground-truth (GT) frequencies exhibited degraded predictive performance in models trained on augmented data. Conversely, this indicates that the no aug model yields poor predictive power for molecules dominated by high GT frequencies; incorporating augmented data helps mitigate underestimation tendencies primarily in the high GT frequency regime. However, because this analysis is limited to a small number of cases, generalization is difficult, and further systematic analysis will be required to draw definitive conclusions.

\subsection{Training cost comparison}
Table~\ref{tab:comput_complex} compares the per-structure, per-step training cost across three pathways that supply curvature information to E-F models. Conventional methods that train using the model's own force Jacobian ($\partial_x \mathbf{F}$) via Hessian supervision (full Hessian loss and row sampling) must recompute curvature within the autograd graph at every training step, requiring vector-Jacobian product (VJP) operations proportional to either the total Cartesian degrees of freedom ($3N$) or the number of sampled rows ($s \ll 3N$). This incurs $3N$ additional backward passes under full Hessian supervision and $s$ additional backward passes under row sampling, introducing substantial memory overheads of $\mathcal{O}(N^2)$ and $\mathcal{O}(sN)$, respectively, to retain activation values prior to parameter updates.

In contrast, our framework leverages fixed DFT Hessians precomputed in the dataset as an external curvature prior. Because augmentation operations are performed entirely outside the autograd graph, they are not recomputed during training. Per-structure computational cost is restricted to a single matrix-vector product for UniAug, with ModeAug requiring only one additional eigendecomposition. Consequently, our approach achieves zero additional backward passes ($0$) and zero backward memory footprint ($0$), perfectly maintaining the computational complexity of the training loop at standard $E\text{--}F$ training levels.

\begin{figure}[htbp]
  \centering
  \includegraphics[width=0.98\linewidth]{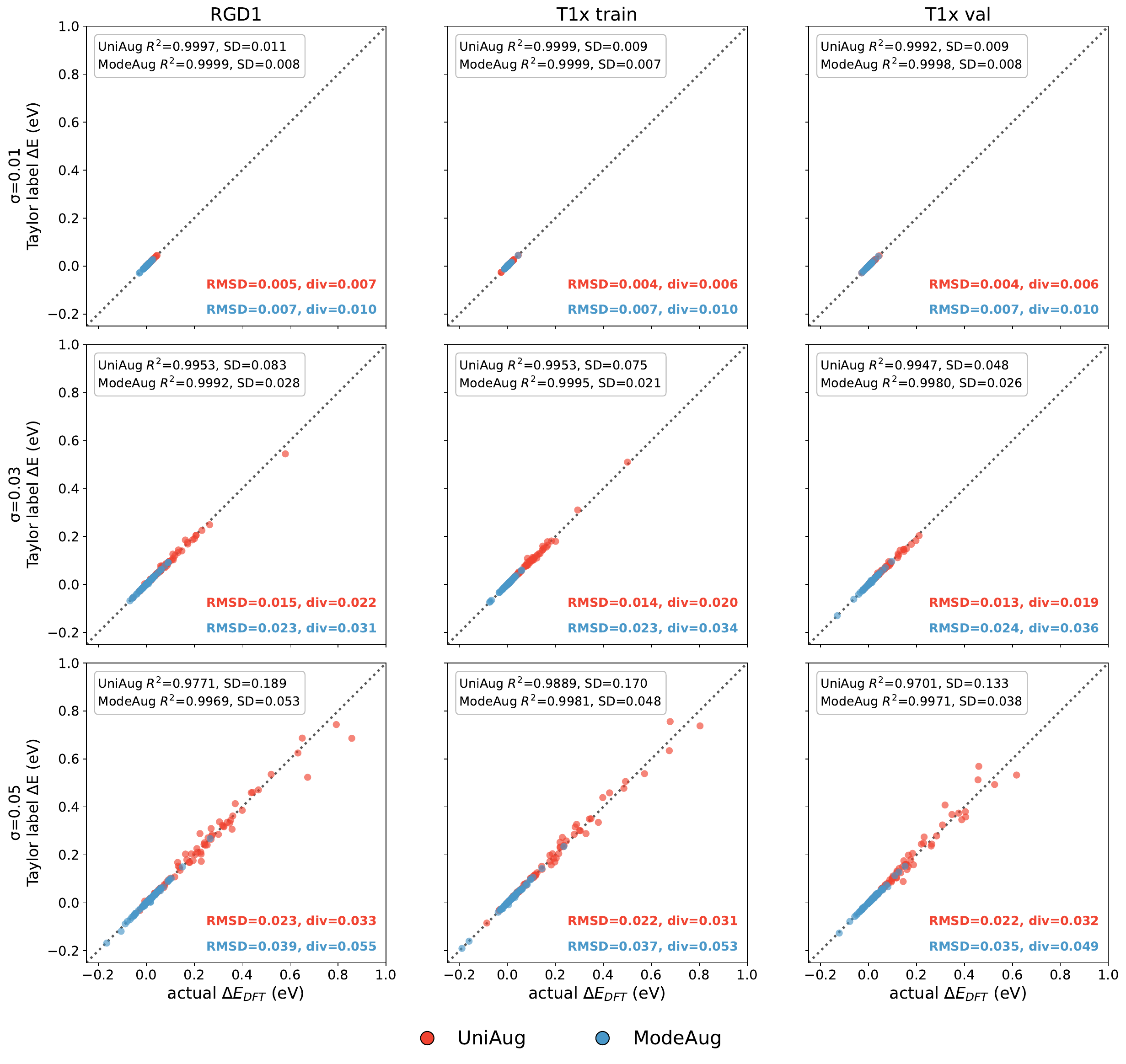}
  \caption{Comparison of the Taylor label $\Delta E$ and the actual $\Delta E_{\mathrm{DFT}}$ for augmented structures. Rows correspond to $\sigma$ (0.01, 0.03, and 0.05), and columns correspond to the HORM subsets (RGD1, Transition-1x train, and Transition-1x val). The dotted line indicates $y = x$, with UniAug in red and ModeAug in blue. Each panel contains 30 molecules $\times$ 3 seeds per method. The box at the top of each panel reports $R^2$ and the standard deviation (SD, in eV) of the actual $\Delta E_{\mathrm{DFT}}$, and the values at the bottom report the mean RMSD from the original structure and the mean RMSD among seeds (div) (in~\AA).}
  \label{fig:taylor_dft}
\end{figure}


\begin{table}[htbp]
  \caption{Validation force MAE on HessianQM9 subsets at the best $\sigma$ range for each subset, where $\langle|\mathbf{F}|\rangle$(meV/\AA) is the mean value of absolute forces in the dataset.}
  \label{tab:hqm9_bestsigma}
  \centering
  \begin{tabular}{llccccc}
    \hline
    Solvents & $\sigma$ range (\AA) & Paper report \citep{williams2025hessian} \textsuperscript{a}  & $\langle|\mathbf{F}|\rangle$ & No aug & UniAug & ModeAug \\
    \hline
    Vacuum  & 0.002--0.02   & 35 & 1.92 & 1.877 & 1.934 & \textbf{1.851} \\
    Toluene & 0.0008--0.02  & 29 & 1.93 & 1.876 & 1.917 & \textbf{1.811} \\
    Water   & 0.0008--0.002 & 28 & 1.38 & \textbf{1.385} & 1.455 & 1.398 \\
    THF     & 0.0008--0.005 & 28 & 1.68 & \textbf{1.686} & 1.781 & 1.729 \\
    \hline
  \end{tabular}
  \par\vspace{3pt}
   {\footnotesize \textsuperscript{a} Not directly comparable, as the paper does not specify its force reduction convention (molecule wise vs atom wise, 18.4 atoms on average).}
\end{table}

\begin{table*}[t]
\centering
\caption{Zero-shot OOD force prediction performance on MD17 for EquiformerV2 models trained on HORM (ts1x-train). Reported values are evaluated via MDsim using the model trained with direct forces predictions.}
\label{tab:md17_stage1}
\footnotesize
\setlength{\tabcolsep}{5pt}
\begin{tabular}{l ccc ccc}
\toprule
& \multicolumn{3}{c}{Force MAE (eV/\AA) $\downarrow$} & \multicolumn{3}{c}{Force cosine $\uparrow$} \\
\cmidrule(lr){2-4} \cmidrule(lr){5-7}
Molecule & No aug & UniAug & ModeAug & No aug & UniAug & ModeAug \\
\midrule
aspirin        & 0.2248 & 0.2275 & \textbf{0.2112} & 0.9557 & \textbf{0.9572} & 0.9571 \\
ethanol        & 0.1641 & 0.1874 & \textbf{0.1627} & \textbf{0.9738} & 0.9725 & 0.9729 \\
naphthalene    & 0.1639 & 0.1729 & \textbf{0.1569} & 0.9804 & 0.9799 & \textbf{0.9809} \\
salicylic acid & 0.2425 & 0.2402 & \textbf{0.2257} & 0.9426 & 0.9451 & \textbf{0.9458} \\
toluene        & 0.1311 & 0.1358 & \textbf{0.1172} & 0.9869 & 0.9888 & \textbf{0.9898} \\
uracil         & 0.2596 & 0.2743 & \textbf{0.2494} & 0.9425 & 0.9423 & \textbf{0.9452} \\
malonaldehyde  & 0.2497 & 0.2812 & \textbf{0.2446} & 0.9493 & 0.9481 & \textbf{0.9502} \\
benzene        & 0.1117 & 0.1063 & \textbf{0.1058} & 0.9780 & 0.9807 & \textbf{0.9823} \\
\midrule
\textbf{MEAN}  & 0.1934 & 0.2032 & \textbf{0.1842} & 0.9637 & 0.9643 & \textbf{0.9655} \\
\bottomrule
\end{tabular}
\end{table*}

\begin{table}[t]
\centering
\caption{Zero-shot OOD MD simulation results on four MD17 molecules (300 ps NVT at 500 K). Metrics include average stability time ($\text{stab}$ in ps), completed trajectories ($\text{\#fin}$ out of 5), and radial distribution function error ($h(r)$ MAE). Best $h(r)$ values among valid trajectories are bolded.}
\label{tab:md17_nh}
\begin{threeparttable}
\small
\setlength{\tabcolsep}{4pt}
\begin{tabular}{l ccc ccc ccc}
\toprule
& \multicolumn{3}{c}{\textbf{No aug.}} & \multicolumn{3}{c}{\textbf{UniAug}} & \multicolumn{3}{c}{\textbf{ModeAug}} \\
\cmidrule(lr){2-4}\cmidrule(lr){5-7}\cmidrule(lr){8-10}
Molecule & stab & \#fin & $h(r)$ & stab & \#fin & $h(r)$ & stab & \#fin & $h(r)$ \\
\midrule
aspirin        & 246.2 & 4/5 & 0.0062 & 300.0 & 5/5 & 0.0040 & 300.0 & 5/5 & \textbf{0.0038} \\
ethanol        & 300.0 & 5/5 & 0.0056 & 300.0 & 5/5 & 0.0047 & 185.3 & 2/5 & \textbf{0.0045} \\
naphthalene    & 300.0 & 5/5 & 0.0120 & 300.0 & 5/5 & 0.0077 & 300.0 & 5/5 & \textbf{0.0060} \\
salicylic acid & \phantom{0}18.3 & 0/5 & (0.0100)$^{\dagger}$ & 300.0 & 5/5 & 0.0069 & 262.3 & 4/5 & \textbf{0.0056} \\
\midrule
\textbf{MEAN}  & 216.1 & 14/20 & 0.0079 & \textbf{300.0} & \textbf{20/20} & 0.0058 & 261.9 & 16/20 & \textbf{0.0050} \\
\bottomrule
\end{tabular}
\vspace{2pt}
\footnotesize
$^{\dagger}$  Excluded from mean and bolding due to zero completed trajectories ($0/5$).
\end{threeparttable}
\end{table}

\begin{table}[htbp]
  \caption{Curvature evaluation metrics (MAE) on ts1x-val (in-domain).}

  \label{tab:curvature_evaluation_ts1x}
  \centering \small \setlength{\tabcolsep}{4pt}
  \begin{tabular}{lcc|ccc|ccc}
    \hline
    & \multicolumn{2}{c|}{Paper report} & \multicolumn{3}{c|}{Direct force}
    & \multicolumn{3}{c}{Autograd force} \\
    \cline{2-3}\cline{4-6}\cline{7-9}
    Metric & E-F-H & E-F & No aug. & UniAug & ModeAug
                         & No aug. & UniAug & ModeAug \\
    \hline
    Hessian      & 0.075 & 2.231  & 1.2248 & 0.6594 & \textbf{0.5387}
                                       & 2.4411 & 0.5677 & \textbf{0.4835} \\
    Eigenvalues          & 0.292 & 20.795 & 9.0690 & 5.1660 & \textbf{3.8176}
                                       & 14.5419 & 3.6846 & \textbf{2.3230} \\
    Asymmetry         & --    & --     & 1.0125 & 0.5172 & \textbf{0.4834}
                                       & 0 & 0 & 0 \\
    $n_\mathrm{imag}$ (GT 0.99) & -- & -- & 2.90 & 1.87 & \textbf{1.65}
                                       & 6.55 & \textbf{1.47} & 1.74 \\
    \hline
  \end{tabular}
\end{table}

\begin{table}[htbp]
  \caption{Curvature evaluation metrics (MAE) on RGD1 (OOD).}
  \label{tab:curvature_evaluation_rgd1}
  \centering
  \small
  \setlength{\tabcolsep}{4pt}
  \begin{tabular}{lcc|ccc|ccc}
    \hline
    & \multicolumn{2}{c|}{Paper report} & \multicolumn{3}{c|}{Direct force}
    & \multicolumn{3}{c}{Autograd force} \\
    \cline{2-3}\cline{4-6}\cline{7-9}
    Metric & E-F-H & E-F & No aug. & UniAug & ModeAug
                         & No aug. & UniAug & ModeAug \\
    \hline
    Hessian       & 0.092 & 1.224  & 0.4149 & \textbf{0.2486} & 0.2530
                                        & 3.8554 & \textbf{0.5807} & 0.7033 \\
    Eigenvalues           & 0.292 & 10.300 & 2.1088 & \textbf{1.2585} & 1.4207
                                        & 25.8133 & \textbf{3.3831} & 4.4429 \\
    Asymmetry          & --    & --     & 0.487  & 0.290  & \textbf{0.263}
                                        & 0      & 0      & 0 \\
    $n_\mathrm{imag}$ (GT 3.86) & -- & -- & 4.80 & 4.58 & \textbf{4.58}
                                        & 10.43  & \textbf{5.11} & 5.19 \\
    \hline
  \end{tabular}
\end{table}

\newpage

\begin{table}[htbp]
    \caption{Per-structure, per-step training cost of three routes to supplying curvature to an E--F model. $N$: number of atoms, $3N$: Cartesian degrees of freedom. Row sampling supervises $s \ll 3N$ rows. UniAug requires one matrix--vector product per structure and ModeAug one additional eigendecomposition.}
  \label{tab:comput_complex}
  \centering
  \renewcommand{\arraystretch}{1.3}
  \small
  \begin{tabular}{l l l l}
    \toprule
     & \textbf{UniAug \& ModeAug} & \textbf{Row sampling}~\citep{cui2025large} & \textbf{Full Hessian loss} \\
    \midrule
    Curvature source     & DFT Hessian (data)      & Model $\partial_x \mathbf{F}$ & Model $\partial_x \mathbf{F}$ \\
    In autograd graph    & No                      & Yes                           & Yes \\
    Recomputed each step & No                      & Yes                           & Yes \\
    Curvature operation  & mat--vec ($+$ eigh)     & VJP $\times s$                & VJP $\times 3N$ \\
    Backward passes      & 0                       & $s$                           & $3N$ \\
    Extra memory in backprop         & 0                       & $O(sN)$                       & $O(N^2)$ \\
    \bottomrule
  \end{tabular}
\end{table}

\section{Discussion}

\paragraph{Fidelity of Augmented Data and Contribution to Predictive Performance}
In this study, we proposed two Taylor-expansion-based data augmentation schemes: ModeAug and UniAug. A t-SNE projection of the displacement vectors confirmed that these two schemes occupy distinct regions, complementarily capturing both physically constrained near-equilibrium vibrations and broader off-equilibrium regions. In addition, we verified the chemical validity of our method by showing that energies derived from Taylor expansions closely align with ground-truth DFT calculations. Importantly, the energy and force shifts predicted on these augmented geometries agree with reference DFT calculations, demonstrating that Taylor-based augmentation enriches the PES without introducing unphysical artifacts. This expanded coverage reduces the equilibrium bias of standard relaxed datasets and improves accuracy on equilibrium and transition state benchmarks as well as in zero shot MD evaluations. Our methodology is particularly effective for nonequilibrium molecules and in data-limited scenarios.

\paragraph{Computational Efficiency and Flexibility of Data Generation}
Data augmentation leveraging Taylor expansions imposes minimal computational and memory overhead across preprocessing, training, and inference. Because our approach operates purely as a data-level intervention rather than incorporating Hessian terms directly into the loss function, the backpropagation graph remains identical to standard energy-and-force training. This design offers two major practical advantages. First, it renders our methodology completely model-agnostic and seamlessly compatible with any standard MLIP architecture. Second, structures can either be pre-generated offline to avoid runtime latency or synthesized on-the-fly to avoid storage overhead. More broadly, this plug-and-play design should allow existing pretrained MLIP foundation models to be fine-tuned without modifying their network architectures or loss formulations, although we leave this direction to future work.

\paragraph{Characteristics of Augmentation Modes and Practical Guidance}

Both augmentation modes improved on the unaugmented baseline where reference forces are large, but the advantage shifted between the two schemes depending on the force scale and the target property of each benchmark. On the HORM dataset, UniAug proved more advantageous in domain, whereas ModeAug yielded superior performance out of domain under direct force models. In contrast, when evaluating the accuracy of Hessians and eigenvalues derived from predicted forces, ModeAug showed a slight superiority in domain, while UniAug marginally outperformed ModeAug in OOD evaluation. This implies that minimizing force prediction errors does not necessarily guarantee a corresponding accuracy in Hessian reconstruction. On the equilibrium dataset HessianQM9, ModeAug improved over the baseline in vacuum and toluene but not in water and THF, the two environments with the smallest force scales, where no scheme surpassed the unaugmented model. In MD simulations, ModeAug achieved higher fidelity in recovering interatomic pair distance distributions, whereas UniAug exhibited superior stability in trajectory completion rates.

The force distributions of original and augmented molecules across datasets offer key insights into the mechanisms driving these performance discrepancies. Examining the energy and force shift distributions relative to the reference structures in HessianQM9 suggests that augmentation modes capable of encompassing the original distribution while covering broader regions are more effective for model enhancement. Because the augmentation scale ($\sigma$) also directly governs the distribution of synthesized configurations, careful consideration of these distributional characteristics when selecting the appropriate mode and scale is recommended for new applications.

\paragraph{Limitations}
We explicitly clarify several limitations of our proposed methodology and analytical framework. First, the optimal augmentation mode and scale ($\sigma$) must be determined prior to training for each benchmark dataset. Although the optimal choice is closely linked to the target energy distribution, relying on a somewhat heuristic inspection of energy shift distributions between original and augmented configurations remains a practical limitation. 

Second, performance gains diminish for equilibrium configurations where force magnitudes approach zero. This represents an inherent constraint of our approach, as reference force vectors themselves serve as essential ingredients for label generation. However, this characteristic also implies that our augmentation framework is particularly well-suited for off-equilibrium configurations where reference labels are sparse. 

Third, because our approach relies on local Taylor expansions, it is fundamentally designed to enrich data density within small displacement regimes. Consequently, our methodology is not intended to model global potential energy surfaces across arbitrarily large conformational changes, imposing limitations in describing potential energy profiles along dihedral rotations or accurately estimating global vibrational modes across entire molecular frameworks.

\section{Conclusion}

In this study, we proposed two data augmentation schemes, UniAug and ModeAug, that utilize molecular Hessians as a source for data generation. By leveraging second-order Taylor expansions, our method generates novel configurations within the local neighborhood of a reference structure, along with their corresponding energy and force labels from a single Hessian, all without requiring additional quantum chemical calculations. Because curvature information is imparted exclusively through the synthesized labels, this framework leaves the model architecture and objective function unchanged while avoiding extra computational overhead during training.

Across diverse benchmarks, including HORM (encompassing reaction paths and transition states), HessianQM9 (composed of equilibrium configurations), and MD17 (evaluating long-time dynamics), both augmentation schemes enhanced predictive accuracy wherever reference forces carry a substantial signal, while the gain diminished for equilibrium structures with near-zero forces. Crucially, our approach substantially recovers the potential energy surface curvature that models trained strictly on energies and forces fail to capture. The two augmentation modes also exhibited complementary strengths across evaluation metrics. ModeAug proved advantageous for in-domain curvature reconstruction and for single-point force accuracy and structural fidelity in zero-shot dynamics, whereas UniAug yielded better curvature metrics out of domain and completed every rollout trajectory. Overall, these findings offer practical guidelines and a computationally efficient route for leveraging curvature information in machine learning potentials when Hessian data is readily available.

\section*{Data and Software Availability}
All datasets analyzed in this study are publicly available. HORM, which includes the Transition1x and RGD1 configurations used for training and evaluation, is available at \url{https://doi.org/10.5281/zenodo.17217897}. HessianQM9 is available at \url{https://doi.org/10.6084/m9.figshare.26363959}. The MD17 trajectories including the data splits and evaluation protocol of the MDsim benchmark are available at \url{https://doi.org/10.6084/m9.figshare.21331245}. Code for the UniAug and ModeAug augmentation schemes is available at \url{https://github.com/fromjade/HessianAug}.

\section*{Author Information}
This work was carried out by the authors in their personal capacity, independently of and without support from their employers. The authors declare no competing financial interest.

\begin{suppinfo}
Descriptions and evaluation protocols of all benchmark datasets including molecular dynamics simulation settings, implementation details of the EquiformerV2 backbone and augmentation hyperparameters, definition of the bond length deviation metric, effect of the local frame detach on autograd force evaluation, construction of the displacement descriptor for t-SNE visualization, and calculation of stiff mode frequencies for the case studies (PDF).
\end{suppinfo}

\section*{Acknowledgements}
This work received no external funding, institutional support, or dedicated computational resources.


During the preparation of this work, the authors used Claude Sonnet and Claude Opus (Anthropic) only to proofread the manuscript and to assist with writing code. After using these tools, the authors reviewed, verified, and edited all generated text and code as needed, and take full responsibility for the content of this publication.

\nocite{rupp2012fast,montavon2013machine,ramakrishnan2015electronic,ramakrishnan2014quantum,marenich2009universal}
\bibliography{references}

\clearpage
\setcounter{page}{1}
\setcounter{figure}{0}
\setcounter{table}{0}
\setcounter{equation}{0}
\renewcommand{\thepage}{S\arabic{page}}
\renewcommand{\thefigure}{S\arabic{figure}}
\renewcommand{\thetable}{S\arabic{table}}
\renewcommand{\theequation}{S\arabic{equation}}

\section*{Supporting Information}

\subsection{Details of datasets and evaluation protocols}

\subsubsection{HORM}

HORM is a benchmark dataset constructed from two reaction repositories, Transition-1x (hereafter \texttt{ts1x})~\citep{schreiner2022transition1x} and RGD1~\citep{zhao2023comprehensive}, which together cover extensive chemical reaction pathways and transition states. In addition to DFT-calculated energies and forces at the $\omega$B97X/6-31G* level, HORM provides full reference Hessian matrices.
Unlike traditional molecular databases—such as QM7~\citep{rupp2012fast}, QM7b~\citep{montavon2013machine}, QM8~\citep{ramakrishnan2015electronic}, and QM9~\citep{ramakrishnan2014quantum}—which predominantly target relaxed geometries with vanishingly small atomic forces, HORM captures non-equilibrium structures along reaction coordinates characterized by substantial, non-negligible forces. In these high-force, far-from-equilibrium regimes, Hessian-based augmentation effectively leverages the rich PES curvature, yielding spatially diverse and physically faithful synthetic perturbations.

Model evaluation follows the official protocols established by the benchmark creators. The training set (\texttt{ts1x-train}) and in-domain validation set (\texttt{ts1x-val}) originate from non-overlapping reaction pathways within the original Transition-1x dataset~\citep{schreiner2022transition1x}. For out-of-distribution (OOD) evaluation, RGD1 serves as a challenging 60k test set featuring reactions where up to two bonds form and break simultaneously, representing a distinct chemical and structural domain from \texttt{ts1x}. Models trained exclusively on \texttt{ts1x-train} are applied directly to RGD1 without fine-tuning to evaluate zero-shot OOD generalization. All model architectures, random seeds, and training hyperparameters are kept strictly identical, isolating the augmentation strategy (\texttt{No aug}, \texttt{UniAug}, or \texttt{ModeAug}) as the sole variable.

All models are optimized using the standard joint loss function $\mathcal{L}$, defined as the weighted sum of energy and force Mean Absolute Error (MAE) terms:
\begin{equation}
  \mathcal{L} = \alpha\,\mathcal{L}_E + \beta\,\mathcal{L}_\mathbf{F},
  \qquad
  \mathcal{L}_E = \frac{1}{B}\sum_{m=1}^{B}\bigl|\hat{E}_m - E_m\bigr|,
  \qquad
  \mathcal{L}_F = \frac{1}{N_{\text{total}}}\sum_{i=1}^{N_{\text{total}}}\bigl\lVert\hat{\mathbf{F}}_i - \mathbf{F}_i\bigr\rVert_2,
  \label{eq:ef_terms}
\end{equation}
where $\alpha$ and $\beta$ are weighting coefficients, $B$ is the batch size, and $N_{\text{total}}$ is the total number of atoms in the batch.

\subsubsection{HessianQM9}

HessianQM9~\citep{williams2025hessian} is a database providing numerical Hessians for equilibrium geometries, comprising $41{,}645$ ground-state organic molecules selected from the $133{,}885$ structures in QM9~\citep{ramakrishnan2014quantum}. To ensure structural heterogeneity rather than arbitrary subsampling, these candidates were selected via UMAP-based dimensionality reduction paired with farthest-point sampling to maximize configurational diversity. For each molecule, HessianQM9 supplies the second-derivative matrix alongside DFT-calculated energies and forces at the $\omega$B97X/6-31G* level. Unlike first-order derivatives (forces), which can be evaluated analytically via the Hellmann--Feynman theorem, the Hessians in HessianQM9 were computed via finite differences.

While conventional molecular databases focus exclusively on vacuum potential energy surfaces, HessianQM9 incorporates environmental effects using the SMD implicit solvation model~\citep{marenich2009universal}. It provides Hessians across three distinct solvent environments—water ($\epsilon_r = 80.0$), tetrahydrofuran (THF, $\epsilon_r = 7.6$), and toluene ($\epsilon_r = 2.4$)—in addition to the gas phase.

Because the structures represent energy minima, the reference forces are near zero. Across the four subsets (vacuum and three solvent phases), the root-mean-square (RMS) of force components ranges from $2.3 \times 10^{-3}$ to $3.1 \times 10^{-3}\,\mathrm{eV/\text{\AA}}$ (e.g., $3.0 \times 10^{-3}\,\mathrm{eV/\text{\AA}}$ in vacuum and $2.3 \times 10^{-3}\,\mathrm{eV/\text{\AA}}$ in water), with mean component values on the order of $10^{-8}\,\mathrm{eV/\text{\AA}}$. These values are roughly two orders of magnitude smaller than those in off-equilibrium datasets such as \texttt{ts1x} ($0.36\,\mathrm{eV/\text{\AA}}$). The force labels in HessianQM9 thus reflect numerical convergence residuals from geometry optimization rather than meaningful physical force signals.

Consequently, HessianQM9 provides a complementary testbed to HORM, probing whether curvature information can be effectively injected in a regime where reference forces offer negligible supervisory signal. Models on HessianQM9 are trained exclusively using the direct force training approach under the joint energy-force objective defined in Eq.~\eqref{eq:ef_terms}. To ensure consistent loss weighting across augmentation schemes, force target normalization is computed independently for each arm based on the empirical force distribution generated by its respective perturbation distribution.

\subsubsection{MD17 and Long-Time Dynamics Evaluation}

While HORM and HessianQM9 assess local single-point prediction accuracy, MD17 evaluates the long-term dynamical stability of MLIPs. Comprising \textit{ab initio} molecular dynamics (AIMD) trajectories for eight small organic molecules, MD17~\citep{chmiela2017machine} serves as a standard benchmark for force prediction. However, as demonstrated by Fu et al.~\citep{fu2023forces}, low single-point force errors do not necessarily guarantee physical simulation fidelity, necessitating autoregressive rollout simulations to evaluate whether a learned potential yields physically valid dynamic trajectories.

Because MD17 lacks ground-truth Hessian matrices and uses a different level of electronic structure theory than our training set ($\omega$B97X/6-31G*), it is excluded from the model training process. Instead, models trained on HORM (\texttt{ts1x-train}) are evaluated directly on MD17 without any fine-tuning, thereby probing long-term dynamical behavior under a strict \textbf{zero-shot transfer} setting.

\paragraph{Simulation Protocol}
Initial atomic velocities are drawn from a Maxwell--Boltzmann distribution at $500\,\mathrm{K}$, and atomic forces are evaluated analytically via automatic differentiation of the predicted potential energy surface ($\mathbf{F} = -\nabla_{\mathbf{X}} E$). To compensate for ASE's default assignment of $3N$ degrees of freedom following center-of-mass (COM) momentum removal (which reduces active physical modes to $3N-3$), the thermostat target temperature is set to $T_{\mathrm{set}} = T \cdot (3N-3)/3N$. This protocol maintains the measured internal temperature strictly within the range of $494\,\mathrm{K}$ to $518\,\mathrm{K}$ across all production runs.

\paragraph{Evaluation Metrics}
GPU non-determinism introduces approximately $0.07\%$ relative variation in force magnitude, which renders individual breakdown times stochastic and motivates the use of the completion count as the primary stability index. The radial distribution histogram $\langle\hat{h}(r)\rangle$ is constructed from all pairwise distances ($0$--$8\,\mathrm{\AA}$, bin width $0.05\,\mathrm{\AA}$, sampled every 20 steps).

\subsection{Implementation Details}
\label{sec:implementation}

We selected EquiformerV2~\citep{liao2024equiformerv2} as our primary backbone architecture, which is $SO(3)$-equivariant, demonstrates strong empirical performance across various atomistic tasks, and serves as the baseline model in the HORM benchmark. Following the configuration established in \citet{cui2025large}, the model consists of 4 layers ($l_{\max}=4$, $m_{\max}=2$, 128 spherical channels), comprising approximately $13.9\text{M}$ parameters. Atomic graphs were constructed using a radial cutoff of \texttt{max\_radius} $= 12\,\mathrm{\AA}$ and \texttt{max\_neighbors} $= 20$. Models were optimized using AdamW (weight decay $10^{-3}$, gradient norm clipping at $10.0$) with a linear warmup followed by a cosine decay schedule (minimum learning rate set to $2\%$ of the initial value).

To verify that the benefits of our augmentation are not specific to a single architecture, we additionally trained GotenNet~\citep{aykent2025gotennet}, an efficient model that exploits high degree geometric information while maintaining $E(3)$ equivariance, without irreducible representations or Clebsch Gordan tensor products. We used the small variant ($\text{GotenNet}_{S}$), which has 4 interaction layers, node, edge, and edge refinement dimensions of 256, $L_{\max}=2$, 8 attention heads, and 64 radial basis functions. Atomic graphs were constructed using a radial cutoff of $5.0\,\mathrm{\AA}$.

All training runs were conducted on a single NVIDIA GeForce RTX 3090 GPU ($24\,\mathrm{GiB}$).

The perturbation scale was set to $\sigma = 0.05$\,\AA{} for HORM. For HessianQM9, $\sigma$ was drawn from a per-solvent range, which is reported together with the results in Table~\ref{tab:hqm9_bestsigma}. For UniAug, the fraction of displaced atoms was $f = 0.125$ on HORM and $f = 0.25$ on HessianQM9. ModeAug displaces all atoms through the mode basis. To maintain a strict $1:1$ balance between original and perturbed structures, data augmentation was applied dynamically at the mini-batch level. While implemented on the fly for computational convenience, this pipeline can also be executed as an offline preprocessing step without loss of generality. We emphasize that this dynamic sampling scheme is an implementation choice for training convenience. 
Across all benchmark datasets, model architectures, and experimental conditions (including baseline and augmented models), training was governed by the identical joint energy-force loss objective defined in Eq.~\eqref{eq:ef_terms}, ensuring that any observed performance gains stem solely from data-level augmentation.

\subsection{t-SNE Descriptor Construction}
\label{sec:tsne-descriptor}


The t-SNE visualization of augmentation displacements in the main text compares displacement fields across molecules with different numbers of atoms $N$, so the raw displacement $\boldsymbol{\delta}\mathbf{X} = \mathbf{X}_{\mathrm{displaced}} - \mathbf{X}_{\mathrm{original}}$ has a different shape for every molecule. 
To enable cross-molecule comparison, each displacement field was converted into a rotation- and permutation-invariant fixed-length descriptor comprising (i) the sorted quantile curve of per-atom displacement magnitudes, (ii) the participation ratio, and (iii) the normalized eigenvalues of the displacement-direction tensor. 
After feature-wise standardization, two-dimensional t-SNE was applied separately to each data source with perplexity $30$, PCA initialization, and $2000$ iterations. Because t-SNE preserves only local structure, comparisons are valid only within each panel.

The energy excitation used for the color scale is defined as $\Delta E = E(\mathrm{displaced}) - E(\mathrm{original})$ and was computed from single-point energies of an EquiformerV2 model trained from scratch without augmentation.
A color scale shared across the four (method $\times$ $\sigma$) categories was clipped at the 98th percentile to suppress the influence of a few high-energy outliers; the histograms above each colorbar pool all sources and mark the $0.1$\,eV threshold together with the fraction of points exceeding it.

\subsection{Bond Length Deviation Metric}
\label{sec:bond-deviation}

To quantify how much augmentation perturbs the local molecular structure, we measured the relative deviation $|\Delta r / r|$ of each bond.

\paragraph{Bond Definition}
The bond list was determined only once based on distance criteria from the original geometry before applying displacements. For atomic pairs $(i,j)$ with $i < j$, a pair was considered a bond if it satisfied
\begin{equation}
  r_{ij} < R_i^{\mathrm{cov}} + R_j^{\mathrm{cov}} + \delta,
  \qquad \delta = 0.4~\text{\AA},
  \label{eq:bond-criterion}
\end{equation}
where $R^{\mathrm{cov}}$ denotes the covalent radius (H $0.31$, C $0.76$, N $0.71$, O $0.66$, and F $0.57~\text{\AA}$). The resulting cutoffs for element pairs are $1.92$ for C--C, $1.87$ for C--N, $1.82$ for C--O, $1.47$ for C--H, $1.37$ for O--H, and $1.02~\text{\AA}$ for H--H.

\paragraph{Metric}
In augmented structures, bonds were not redetected, and the relative deviation
\begin{equation}
  \frac{\Delta r}{r}
  =
  \frac{r_{ij}^{\mathrm{aug}} - r_{ij}^{\mathrm{orig}}}{r_{ij}^{\mathrm{orig}}}
  \label{eq:bond-deviation}
\end{equation}
was calculated over the fixed set of atomic pair indices from Equation~\eqref{eq:bond-criterion}. Because the distributions were symmetric around zero across all three conditions, we reported the folded magnitudes $|\Delta r / r|$. Applying the identical bond list to both UniAug and ModeAug allowed us to evaluate the two augmentation schemes through paired comparisons.

\paragraph{Validity and Limitations}
The distance cutoff in Equation~\eqref{eq:bond-criterion} is a geometric heuristic that does not account for bond order, valence, or electronic structure. However, in the organic molecules containing H, C, N, O, and F studied here, the gap between bonding distances and nonbonding nearest neighbor distances is sufficiently wide. For example, C--C bonds at $1.20$ to $1.54~\text{\AA}$ stand well apart from $1,3$-C$\cdots$C distances near $2.5~\text{\AA}$, C--H at $1.09~\text{\AA}$ stands apart from $1,3$-H$\cdots$C distances near $2.1~\text{\AA}$, geminal H$\cdots$H distances are around $1.78~\text{\AA}$, and hydrogen bonded O$\cdots$H distances at $1.8$ to $2.0~\text{\AA}$ lie outside the cutoff, so this criterion reliably separates bonds.

On the other hand, in transition states, the distances of partial bonds being formed or broken lie near the cutoff. For instance, C$\cdots$H distances of $1.3$ to $1.4~\text{\AA}$ in hydrogen transfer transition states include both partial bonds, while forming C$\cdots$C distances of $2.0$ to $2.4~\text{\AA}$ straddle the $1.92~\text{\AA}$ cutoff, meaning this definition does not align perfectly with chemical bond lists.

However, because such misclassifications apply equally to both augmentation conditions, our relative comparison between the two schemes, specifically the finding that ModeAug produces a shorter tail in the $|\Delta r / r|$ distribution, remains insensitive to the details of the bond definition. On the other hand, this limitation applies to absolute chemical interpretations, such as treating a specific threshold as bond cleavage. Therefore, this metric was used solely for comparison between augmentation schemes.

\subsection{Effect of the Local Frame Detach on Autograd Force Evaluation}

EquiformerV2 is designed around direct force prediction, and its local rotation frame $R(\mathbf{x})$ serves as a geometric preprocessing step that reduces SO(3) convolutions to SO(2). Because this frame does not depend on the model parameters, the upstream implementation applies \texttt{.detach()} along this path to reduce the cost of building the computational graph. Under this setting the force obtained by autograd is not the true gradient of $E = f(\mathbf{x}, R(\mathbf{x}))$ but the gradient of a proxy function in which the frame is held fixed, and the model minimizes the force loss through this same path. We therefore followed the original design intent and kept the detach in place during training. When the detach was removed at evaluation time after training, however, we found that the error metrics for curvature changed appreciably, in some cases decreasing. We report this observation not to argue for or against removing the detach, nor to extend it into any further interpretation, but to note that it is one variable that affects the evaluation results.

\subsection{Stiff Mode Frequency Calculation for Case Studies}

Predicted and reference Hessians were mass weighted as $\mathbf{D} = \mathbf{M}^{-1/2} \mathbf{H} \mathbf{M}^{-1/2}$, where $\mathbf{M}$ is the diagonal matrix of atomic masses repeated over the three Cartesian components, and diagonalized to obtain eigenvalues $\lambda_i$. Since the predicted Hessian from the direct force head is not strictly symmetric, only its symmetric part $\tfrac{1}{2}(\mathbf{H}+\mathbf{H}^{\mathsf{T}})$ was used prior to diagonalization. The six eigenvalues with smallest $|\lambda_i|$, corresponding to translational and rotational modes, were discarded, and signed vibrational frequencies were computed from the remaining $3N-6$ modes as
\begin{equation}
\nu_i = \operatorname{sign}(\lambda_i)\,\sqrt{|\lambda_i|}\;\times\;521.47\ \text{cm}^{-1},
\end{equation}
with $\lambda_i < 0$ yielding imaginary frequencies reported as negative values. Each normal mode was assigned to a stiff or soft band according to its reference frequency, with $|\nu_i^{\text{ref}}| \geq 400\ \text{cm}^{-1}$ defining the stiff band and $|\nu_i^{\text{ref}}| < 400\ \text{cm}^{-1}$ defining the soft band. The stiff mode error was then defined as the mean absolute error between predicted and reference frequencies averaged only over modes falling in the stiff band.

\newpage

\end{document}